\documentclass[10pt,twocolumn,letterpaper]{article}

\usepackage{cvpr}

\usepackage[algo2e,ruled,noend]{algorithm2e}

\usepackage{CJKutf8}
\usepackage{booktabs}
\usepackage{colortbl}
\usepackage{subcaption}
\usepackage{multirow}
\usepackage{makecell}
\usepackage{algorithm}
\usepackage{algorithmic}
\usepackage{marvosym}
\usepackage[accsupp]{axessibility}

\definecolor{cvprblue}{rgb}{0.21,0.49,0.74}
\usepackage[pagebackref,breaklinks,colorlinks,allcolors=cvprblue]{hyperref}

\def\confName{CVPR}
\def\confYear{2025}

\newcommand{\promptbox}[1]{
\begin{center}
\setlength{\fboxsep}{6pt} 
\fbox{
    \parbox{0.95\linewidth}{
        \renewcommand{\baselinestretch}{0.8}\selectfont 
        \vspace{3pt} 
        \texttt{\scriptsize #1} 
        \vspace{3pt}
    }
}
\end{center}}

\newcommand\blfootnote[1]{%
\begingroup
\renewcommand\thefootnote{}\footnote{#1}%
\addtocounter{footnote}{-1}%
\endgroup
}

\def\modelname{Docopilot\xspace}
\def\dataname{Doc-750K\xspace}

\title{Docopilot: Improving Multimodal Models for Document-Level Understanding}

\author{\scalebox{0.91}{Yuchen Duan$^{1,2*}$, Zhe Chen$^{3, 1*}$, Yusong Hu$^{4,1*}$, Weiyun Wang$^{*5,1}$, Shenglong Ye$^1$, Botian Shi$^1$,} \\
\vspace{4px}
\scalebox{0.91}{~Lewei Lu$^7$, Qibin Hou$^4$, Tong Lu$^{3,1}$, Hongsheng Li$^{2,1}$, Jifeng Dai$^{6,1}$, Wenhai Wang$^{2,1}$\textsuperscript{\Letter}} \\
\scalebox{0.91}{~$^1$Shanghai AI Laboratory,~$^2$The Chinese University of Hong Kong, $^3$Nanjing University,} \\
\scalebox{0.91}{~$^4$Nankai University, $^5$Fudan University, $^6$Tsinghua University, $^7$SenseTime Research}
\vspace{-2ex}
}

\begin{document}
\begin{CJK}{UTF8}{gbsn} 
\maketitle

\blfootnote{*\ Equal contribution; \\ \indent ~\Letter\ Corresponding author:~wangwenhai@pjlab.org.cn}

\begin{abstract}
Despite significant progress in multimodal large language models (MLLMs), their performance on complex, multi-page document comprehension remains inadequate, largely due to the lack of high-quality, document-level datasets.
While current retrieval-augmented generation (RAG) methods offer partial solutions, they suffer from issues, such as fragmented retrieval contexts, multi-stage error accumulation, and extra time costs of retrieval. 
In this work, we present a high-quality document-level dataset, \dataname, designed to support in-depth understanding of multimodal documents.
This dataset includes diverse document structures, extensive cross-page dependencies, and real question-answer pairs derived from the original documents.
Building on the dataset, we develop a native multimodal model—\modelname, which can accurately handle document-level dependencies without relying on RAG.
Experiments demonstrate that \modelname achieves superior coherence, accuracy, and efficiency in document understanding tasks and multi-turn interactions, setting a new baseline for document-level multimodal understanding. 
Data, code, and models are released at \url{https://github.com/OpenGVLab/Docopilot}.
\end{abstract}

\vspace{-2ex}
\section{Introduction}
\label{sec:intro}
In recent years, multimodal large language models (MLLMs)~\cite{liu2023improved, chen2023internvl, bai2023qwenvl, wang2024qwen2vl, wang2023allseeing, wang2023cogvlm, gpt4v, team2023gemini, reid2024gemini1_5, yao2024minicpm_v} have rapidly developed, achieving remarkable performance in various visual understanding tasks~\cite{tu2024overview, jiang2024effectiveness}, particularly image-level tasks, such as image captioning~\cite{chen2015cococaption, liu2023llava}, optical character recognition (OCR)~\cite{liu2023ocrbench, singh2019textvqa}, and visual question answering (VQA)~\cite{liu2023mmbench, fu2023mme}.
Despite these advances, current MLLMs still face significant challenges in document-level understanding~\cite{ma2024mmlong, xia2024docgenome, tito2023mpdocvqa}, where models are required to identify and integrate key information across multi-page documents, setting high expectations for their long-context processing capabilities of MLLMs.

\begin{figure}[t!]
    \centering
    {\includegraphics[width=\linewidth]{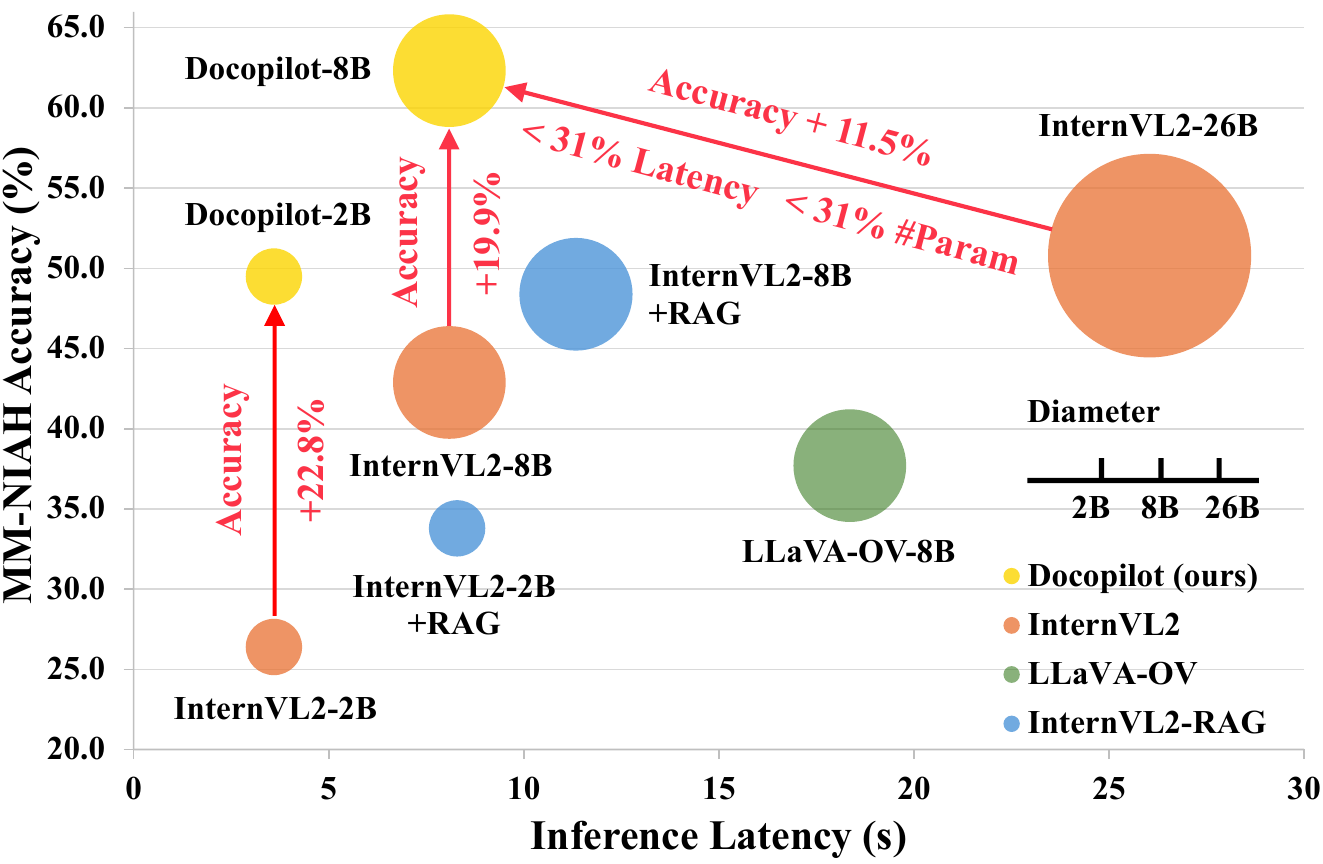}}
    \caption{\textbf{Accuracy \emph{v.s} inference latency on MM-NIAH.} 
    The proposed \modelname-8B shows a notable improvement over baseline models~\cite{InternVL2}, achieving a +19.9\% accuracy gain compared to InternVL2-8B and surpassing InternVL2-26B with less than 31\% of the inference latency. Additionally, \modelname-2B uses fewer parameters (less than 10\%) while exhibiting comparable performance to the 10$\times$ larger InternVL2-26B. These results suggest that our \modelname strikes a reasonable balance between latency, model size, and performance.}
    \label{fig:teaser}
    \vspace{-3mm}
\end{figure}

Current research on long-content understanding primarily focuses on text-only models~\cite{qwen, qwen2.5, cai2024internlm2}, targeting specific retrieval tasks such as ``Needle in a Haystack'' (NIAH)~\cite{an2023leval, LLMTest_NeedleInAHaystack}.
However, existing open-source MLLMs \cite{chen2023internvl, wang2024qwen2vl, li2024llavaonevision, chen2024far, liu2023improved,shi2024eagle,wang2024allseeingv2} are primarily trained on image-level data, lacking the long-context understanding capacity required for document-level understanding.
Retrieval-augmented generation (RAG) methods~\cite{yu2024visrag, cho2024m3docrag,faysse2024colpali,ma2024unifying,sharifymoghaddam2024unirag,shi2023replug} attempt to address this by retrieving key information to fit within the limited context windows of MLLMs, but they still encounter the following challenges in document-level tasks.
(1) \emph{Fragmented Retrieval Contexts}. Retrieved information is fragmented, lacking the overall structure of the document; 
(2) \emph{Multi-Stage Error Accumulation}. Incorrect retrieval results can affect subsequent responses, leading to errors or omissions of critical details, especially in multi-turn or complex tasks;
(3) \emph{Extra Time Costs}. The retrieval step increases the latency of the QA system, limiting the scalability of RAG in time-sensitive scenarios.

To address these problems, two primary challenges need to be considered.
(1) \emph{High-Quality Multimodal Document Dataset}. While extensive datasets \cite{chen2023longlora, yang2023longqlora, zhang2024longreward, zhang2024longcite} exist for long-context, text-only tasks, high-quality document-level question-answering datasets remain scarce. This shortage is largely attributed to the high costs associated with annotation and the lack of streamlined construction pipelines.
(2) \emph{Native Document-Level MLLMs}. Although RAG-based methods \cite{yu2024visrag, cho2024m3docrag, wang2024mmniah} provide some relief, native multimodal models with long-context processing abilities are crucial. However, training native MLLMs specifically for document-level understanding is constrained by current hardware limitations.

In this work, we introduce a new multimodal document dataset that supports document-level understanding tasks. Compared to counterparts \cite{van2023document, laurenccon2024docmatix, tito2023mpdocvqa}, this dataset has the following features:
(1) \emph{Large Scale}. It includes a total of 758K question-answer samples, containing 5.2B text tokens and 3.1M images. It encompasses content from various sources, such as Sci-Hub, Arxiv, and OpenReview, covering a wide range of topics and document layouts.
(2) \emph{High Quality}. Unlike existing datasets that insert irrelevant questions into documents, we collect real, in-depth question-answer pairs and construct single-page and cross-page questions based on document structure. Such high-quality question-answer data accounts for 31.6\% of the dataset. 
(3) \emph{Multimodal}. For the document content, we provide not only the conventional interleaved text-image context but also purely rendered image inputs, catering to the needs of different models.

Building upon this dataset, we developed a native baseline model for document-level multimodal understanding--\modelname.
Unlike existing approaches~\cite{yu2024visrag, cho2024m3docrag, wang2024mmniah} that rely on RAG, our model achieves efficient document-level training and testing through simple engineering optimizations, such as multimodal data packing, Ring Attention \cite{liu2023ring}, and Liger Kernel \cite{dai2024ligerkernel}.
Leveraging the proxy tasks carefully designed within the dataset, \modelname can directly handle long-distance dependencies and cross-page information integration without external retrieval support.
As shown in Figure \ref{fig:teaser}, this approach not only enhances coherence and accuracy compared to RAG methods but also significantly reduces the response time of the entire question-answering system, delivering superior real-time performance in multi-turn interactions.

The main contributions are summarized as follows:

(1) We develop the first large-scale, high-quality dataset for document-level multimodal understanding, consisting of 758K QA pairs from 3 sources, supporting 9 types of proxy tasks. This dataset includes 31.6\% real QA pairs directly extracted from documents.

(2) Based on the dataset, we implement Docopilot, a native MLLM designed for document-level understanding without relying on retrieval mechanisms. 
This approach greatly improves its ability to integrate and comprehend information across multi-page documents.

(3) Through extensive experiments on multiple document-level benchmarks, our method demonstrates performance significantly superior to existing approaches, proving its effectiveness and generality. 
As shown in Figure \ref{fig:teaser}, \modelname-8B achieves a score of 61.8 on MM-NIAH~\cite{wang2024mmniah}, outperforming InternVL2-8B by 19.9 points and surpassing InternVL2-26B with less than 31\% of the latency.
We hope this work could provide a baseline for future advancements in MLLMs for document-level tasks.

\section{Related Work}

\subsection{Multimodal Large Language Models}

Multimodal large language models (MLLMs) have demonstrated impressive capabilities in processing image and text information, opening up new directions for applications such as visual question answering and image captioning.
Early models~\cite{openclip, rebuffi2017learning, li2022blip, chen2023internvl} trained with contrastive learning methods excelled in recognizing and understanding open-world semantics within an image-text matching framework. However, their limited generative abilities restricted their applicability.
To leverage the powerful generation abilities of large language models (LLMs), subsequent works~\cite{li2023blip2, liu2023llava, chen2024far, wang2023allseeing, wang2024allseeingv2,2023interngpt} introduced a connector to align the embedding spaces of vision encoders and LLMs, allowing encoded image embeddings to serve as soft prompts for LLMs. 
Another series of works~\cite{alayrac2022flamingo,li2024omnicorpus,laurenccon2024obelics,zhu2024mmc4} extended LLMs by integrating additional visual experts, reducing reliance on standalone vision encoders.
More recently, models capable of both understanding and generating images have also made notable progress~\cite{tian2024mminterleaved, dong2023dreamllm, sun2023emu, li2023seed}, leveraging the insight that image generation can enhance image understanding.
Despite these advancements, current MLLMs still face challenges with long-context multimodal inputs. For instance, InternVL 2.0~\cite{chen2024far,gao2024mini_internvl} performs optimally within a token range of up to 8192, constraining its effectiveness in document-level applications.

\begin{figure*}[t!]
    \centering
    {\includegraphics[width=\linewidth]{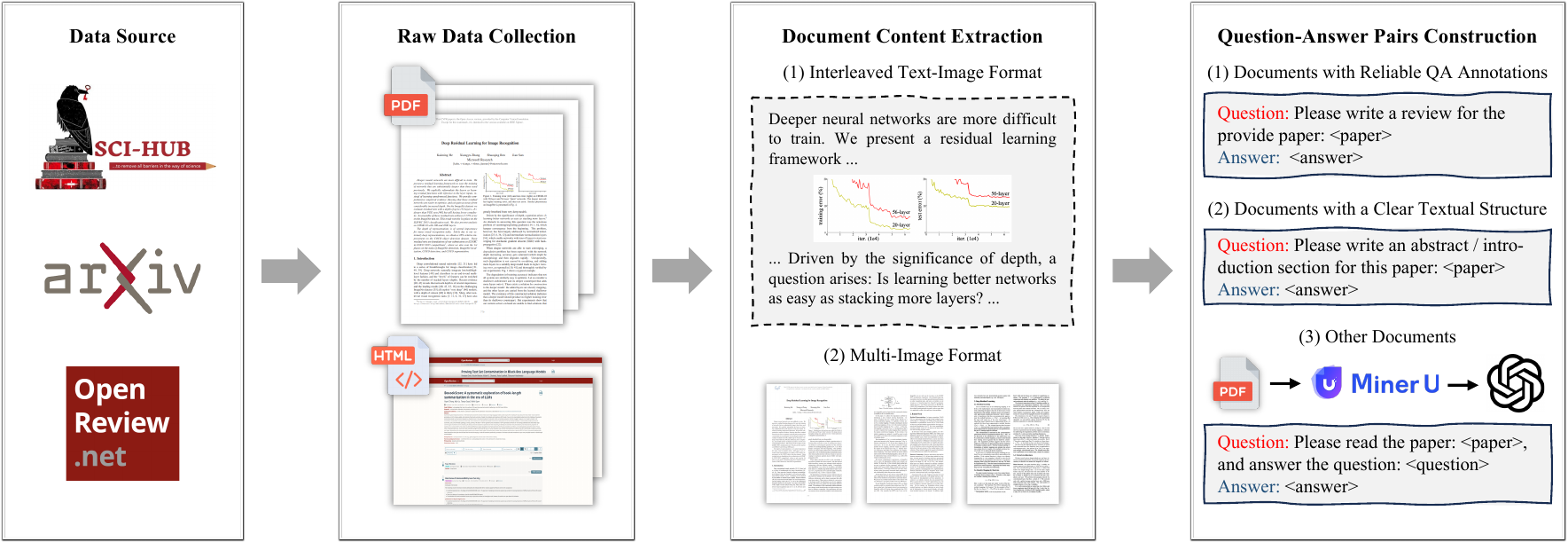}}
    \caption{
    \textbf{Multimodal document dataset generation pipeline.} 
    This pipeline involves three main stages: 
    (1) Raw Data Collection: Documents are gathered from sources like Sci-Hub, arXiv, and OpenReview, available in PDF and HTML formats. 
    (2) Document Content Extraction: Multimodal content is processed in two formats: interleaved text-image format and multi-image format. 
    (3) Question-Answer Pairs Construction: QA pairs are generated based on the document structure or constructed using GPT-4o.
    }
    \label{fig:data_pipeline}
    \vspace{-2mm}
\end{figure*}

\subsection{Document Understanding Models}

Extracting key information from documents is crucial for industries and academic research.
OCR-model-driven methods~\cite{wang2023docllm,appalaraju2024docformerv2,bai2022wukong,tang2023unifying} represent one of the primary technical approaches. These methods extract text, layout, and bounding box information from external systems and integrate it with another model. However, they are prone to error propagation and high processing times due to their reliance on multiple components.
Benefitting from the rapid advancements in LLMs, OCR-free methods have also achieved great progress. Donut~\cite{kim2022ocr} is the first end-to-end training framework based on a Transformer without requiring OCR engines or APIs. 
Subsequent works \cite{wang2023towards,feng2023docpedia,ye2023ureader,wei2023vary,luo2024layoutllm} propose diverse modifications in model architectures and training algorithms.
However, these models are designed for specific tasks and lack general abilities.

\subsection{Long-Context Large Language Models}

With advancements in engineering, architecture, and algorithms, long-context large language models have made substantial progress. Techniques such as Flash Attention \cite{dao2022flashattention,dao2023flashattention} and Ring Attention \cite{liu2023ring} have notably reduced GPU memory usage for training on extended contexts. Additionally, various sparse attention mechanisms—including Shifted Sparse Attention~\cite{chen2023longlora}, Dilated Attention~\cite{ding2023longnet}, and Attention Sinks~\cite{han2023lm,xiao2023efficient}—have enabled efficient scaling to handle larger contexts. 
New positional embedding methods, like ALiBi~\cite{press2021train}, xPOS~\cite{sun2022length}, and RoPE~\cite{su2024roformer}, further enhance the models’ generalization capabilities in length extrapolation.
However, these advancements remain largely confined to natural language processing, and methods to extend the context size of MLLMs are still under-explored.
Another research approach aims to reduce context size by leveraging retrieval augmented generation (RAG)~\cite{yu2024visrag,faysse2024colpali, cho2024m3docrag}, where only the most relevant passages are retrieved and fed into the generation model. However, this retrieval-based approach can disrupt the coherence of the semantic chain, particularly in complex reasoning tasks, due to fragmented information flow.
In this work, we integrate the above engineering techniques into MLLMs and demonstrate that a model fine-tuned on a high-quality, long-context training corpus is a strong baseline, achieving superior performance compared to its RAG counterpart.

\begin{figure*}[t!]
    \centering
    {\includegraphics[width=\linewidth]{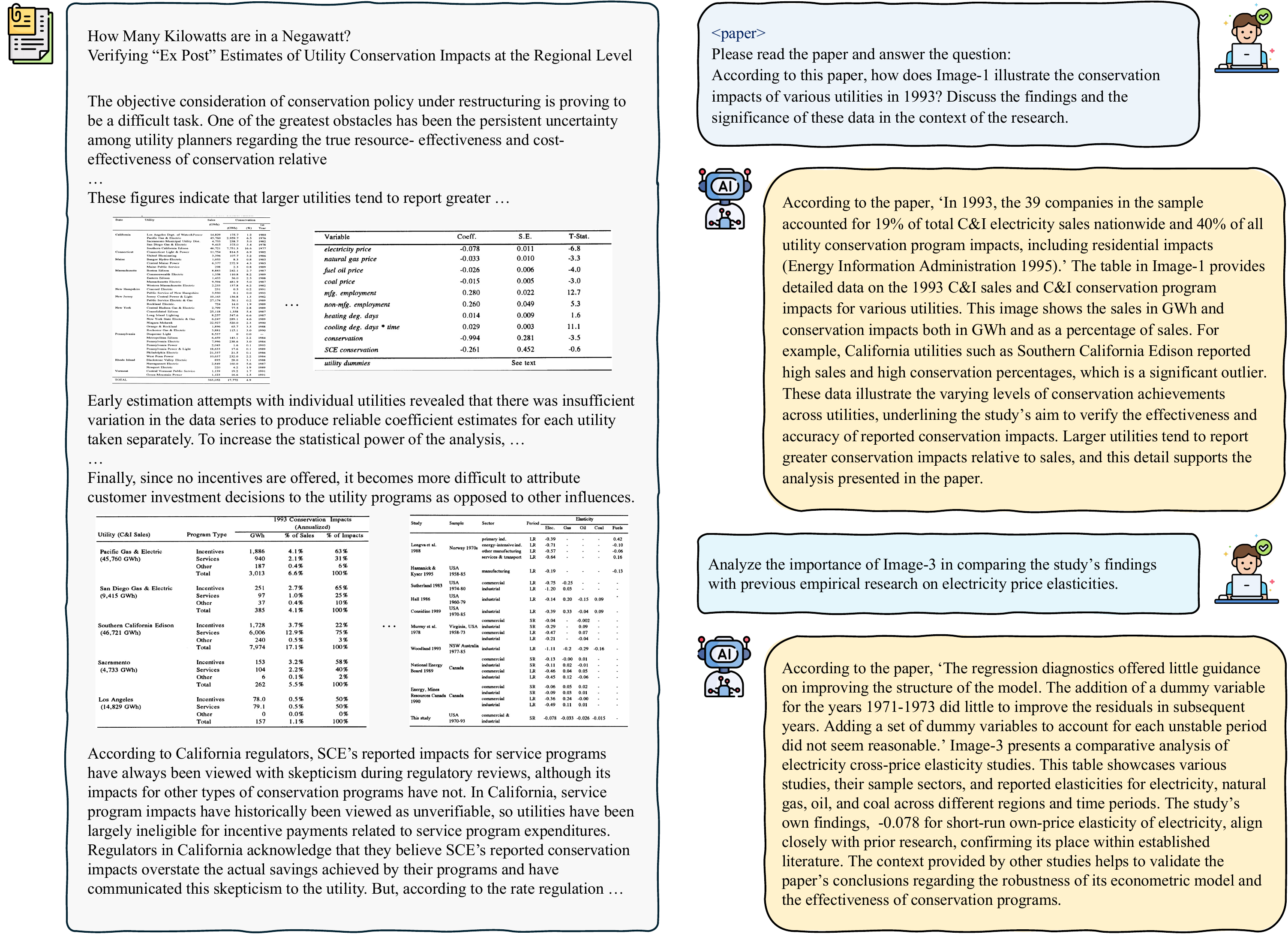}}
    \caption{\textbf{Visualization of an example from {\dataname}.} The left side presents the interleaved text-image format data obtained through Document Content Extraction, while the right side showcases the annotations generated via Question-Answer Pairs Construction.}
    \label{fig:data_example}
    \vspace{-3mm}
\end{figure*}

\section{Multimodal Document Dataset Generation}

In this section, we begin by introducing the details of the data engine.
Following this, we provide a comprehensive overview of the dataset---\dataname.

\newcommand\hys[1]{\textcolor{red}{#1}}
\subsection{Data Engine} 
\label{sec:data_engine}
The data engine operates primarily through two steps: document content extraction and question-answer (QA) pair construction.
Specifically, we first extract multimodal content, including both text and images, from the documents. Based on this extracted content, we then create question-answer pairs. The document content and these constructed pairs are combined to produce conversational-style training data. The format is outlined as:

\begin{center}
\small
\fbox{
\parbox{0.95\linewidth}{
\texttt{Please read the paper: <paper>, and answer the question: <question> Answer: <answer>}
}
}
\end{center}

\noindent Here, the \texttt{<paper>}, \texttt{<question>}, and \texttt{<answer>} are the placeholder for extracted document content, the generated questions and answer, respectively. In the following, we will provide a detailed explanation of the two key steps: document content extraction and QA pair construction.

\newcommand\ysl[1]{\textcolor{blue}{#1}}

\noindent\textbf{Document Content Extraction.}
In practical applications, different documents have varying page layouts and content types, which poses significant challenges for content extraction. To enhance the efficiency of multimodal models, it is necessary to organize documents into a unified format for streamlined processing.
In this work, we process each document into two formats as follows:

(1) \emph{Interleaved Text-Image Format.} Using the document content extractor MinerU~\cite{wang2024mineru}, we segment the document content into interleaved text and image annotations, for example, 
``\texttt{<text>\textbackslash n<image>\textbackslash n<text>\textbackslash n<image>}''
This format captures the document’s textual content, making it easier to construct question-answer pairs.

(2) \emph{Multi-Image Format.} In this format, a document with $n$ pages is rendered as $n$ images, with each image corresponding to a single page. The structure follows the pattern ``\texttt{<image>\textbackslash n<image>\textbackslash n<image>}''. This format preserves the original layout, enabling the model to learn the overall pagination and visual layout of the document.

After processing the document into contexts in interleaved text-image and paginated image formats, we can not only use these contexts for next-token prediction training but also leverage the document's content, hierarchical structure, and layout features to flexibly and precisely generate high-quality question-answer pairs.

\noindent\textbf{Question-Answer Pairs Construction.}
In this step, we create question-and-answer pairs tailored to the source, content features, and formatting structure of each document. This process is divided into the following main categories:

(1) \emph{For documents with reliable QA annotations,} like the review and reply in OpenReview, we extract the QA pairs and organize them into conversation format.

\begin{figure*}[t!]
    \centering
    {\includegraphics[width=\linewidth]{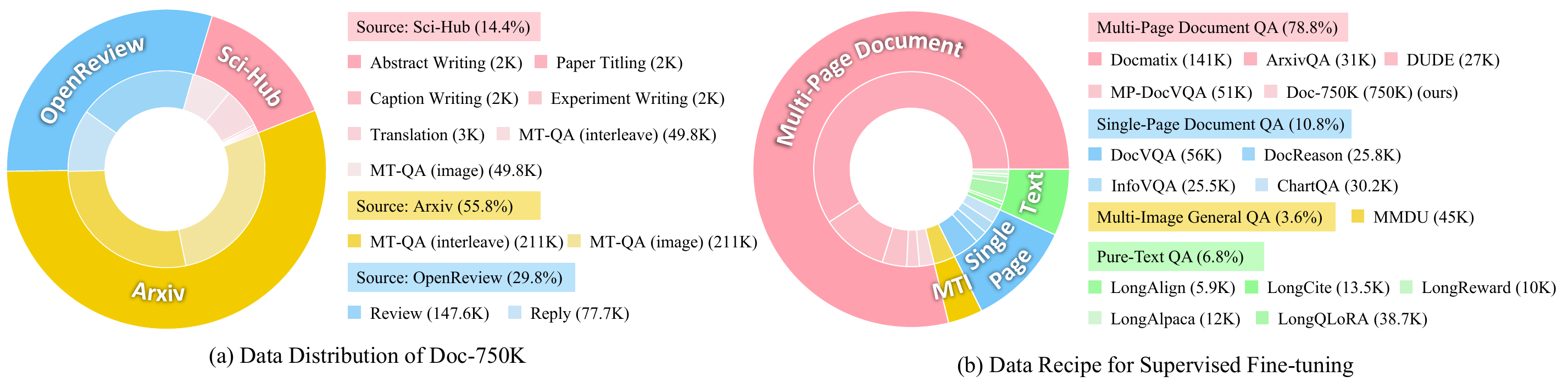}}
    \caption{
        \textbf{Data distribution of our dataset.}
        The outer circle shows the distribution of all data categories and the inner circle shows the distribution of data subsets.
        \textbf{Left:} Data distribution of {\dataname}.
        \textbf{Right:} Data distribution of our complete SFT training dataset. 
        Note that the number reported in the figure represents the number of samples. ``MT'' is short for multi-turn.
    }
    \label{fig:data_distribution}
    \vspace{-3mm}
\end{figure*}

(2) \emph{For documents with a clear textual structure}, such as well-structured papers from Sci-Hub and Arxiv, we convert them to text and segment them, while the model is instructed to generate contents for each segment, including abstracts, experiment descriptions, and captions for figures and tables. The details of each task for structural papers are illustrated in Table~\ref{tab:task_format}. 

(3) \textit{For other documents}, in addition to using them directly for NTP pretraining, we can input text interspersed with images into MLLMs to obtain QA pairs. To ensure high-quality generated data, we use the state-of-the-art model GPT-4o~\cite{gpt4v}.

Through our pipeline, most data has been processed into high-quality document-level question-answering data, while the remaining data is converted to plain text and used for next-token prediction tasks. Our pipelines are meticulously designed to ensure high data quality across all generated context. Each LLM-generated sample is explicitly marked in the metadata as model-generated. Across the entire dataset, only 4.8\% of the data is LLM-generated, reinforcing the overall reliability and quality of the dataset.

\begin{table}[]
\small
\begin{tabular}{lp{4.7cm}}
\toprule
Tasks             & \multicolumn{1}{c}{Questions} \\ 
\midrule
Abstract Writing    &  
Read the full text of the paper and provide a concise summary in the form of an abstract.     
\\
\midrule
Paper Titling      &  Based on the provided abstract or introduction of the research paper, please generate a concise and informative title          \\
\midrule
Caption Writing    &  Give the relative texts of the images or tables, please write a caption for each image or table based on the relative texts provided.   \\
\midrule
Experiment Writing &  Please write the "Experiments" section based on the incomplete research paper provided.       \\
\midrule
Translation        &  Please read the full text of the following research paper and translate the Experiments section into Chinese.         \\
\bottomrule
\end{tabular}
\caption{\textbf{Questions format for different tasks.}
For documents with a clear textual structure, we design several proxy tasks. All tasks leverage the inherent structure of the documents, with answers directly sourced from the original text.
}
\label{tab:task_format}

\vspace{-3mm}

\end{table}

\subsection{Multimodal Document Dataset}

\textbf{Data Source.} 
The composition and distribution of our training data are detailed in Figure~\ref{fig:data_distribution}. Specifically, our dataset predominantly consists of academic papers, which constitute approximately 32.6\% of the total data. The multimodal data, carefully selected to augment our model's learning dimensions, makes up about 88.8\% of our dataset. This strategic distribution is designed to optimize the training process and improve the model's ability to generalize across different types of data inputs.

\noindent\textbf{Dataset Statistics.}
In our {\dataname} dataset, the majority of the data consists of reliably annotated entries, with OpenReview and Arxiv collectively accounting for 75.4\%. The remaining data, sourced from Sci-Hub, is processed using our designed tasks. The overall distribution and number of tasks are shown in Figure~\ref{fig:data_distribution}(a).
Our dataset ultimately consists of 251K conversations, comprising a total of 758K questions. Additional statistical details are provided in Table~\ref{tab:dataset_analysis}. Compared to previous datasets, {\dataname} contains a larger number of images, with an average of four images per conversation segment. Further comparisons with other datasets are shown in Table~\ref{tab:dataset_comparison}.

\begin{table}[]
\small
\centering
\begin{tabular}{lr}
\toprule
Statistics & \multicolumn{1}{c}{Number} \\ \midrule
Total Questions       & 758K   \\
Total Images          & 3.1M   \\
Total Conversations   & 251K   \\ 
\midrule
Multi-Turn Questions  & 87K    \\
Single-Turn Questions & 164K   \\ 
\midrule
Average Text Tokens   & 11245  \\
Average Image Tokens  & 6178   \\ 
\bottomrule
\end{tabular}
\caption{\textbf{Key statistics of the {\dataname} datasets.}
It comprises 758K questions, 3.1M images, and 251K conversations, including 87K multi-turn and 164K single-turn questions. With an average of 11,245 text tokens and 6,178 image tokens, it highlights the dataset’s richness and diversity for multimodal research.
}
\label{tab:dataset_analysis}

\vspace{-3mm}

\end{table}

\subsection{Data Recipe for Supervised Fine-Tuning}

Although \dataname effectively covers multimodal document QA scenarios, using it directly may lead to model over-fitting on a specific document domain. Therefore, we combine it with several open-source datasets to create a mixed dataset for SFT training.
As shown in Figure \ref{fig:data_distribution}(b), these datasets are organized into 4 categories as follows:

 (1) \emph{For multi-page document QA}, \dataname serves as the core dataset, specifically curated to address complex, multi-page document comprehension. Additional datasets such as MP-Docmatix~\cite{laurenccon2024docmatix}, MP-DocVQA~\cite{mathew2021docvqa}, DUDE~\cite{van2023document}, and Taesiri-ArxivQA~\cite{arxivqa}
offer valuable multi-page scenarios requiring inter-page reasoning and contextual retention across sequences. 

 (2) \emph{For multi-image general QA}, MMDU-45K \cite{liu2024mmdu} offers a comprehensive dataset encompassing diverse real-world scenarios, such as natural environments and everyday contexts. It emphasizes multi-turn dialogues and integration of multiple images, supporting the development of systems capable of generating coherent and accurate responses from complex, lengthy inputs.

(3) \emph{For single-page document QA}, We introduce DocVQA~\cite{mathew2021docvqa}, DocReason \cite{ye2023mplugdocowl}, InfoVQA \cite{mathew2022infographicvqa}, and ChartQA \cite{masry2022chartqa} to further enhance the diversity of the SFT dataset.
These datasets focus on individual pages with complex layouts, rich textual information, and, in some cases, graphical data interpretation.

(4) \emph{For pure-text QA}, we add datasets including LongAlpaca \cite{chen2023longlora}, LongAlpaca-16K-Length \cite{chen2023longlora}, LongQLoRA \cite{yang2023longqlora}, LongCite \cite{zhang2024longcite}, LongAlign \cite{bai2024longalign}, and LongReward \cite{zhang2024longreward} to support the assessment of the model's capabilities in QA tasks requiring long-range dependencies.

This expanded dataset provides a balanced foundation for training and evaluating multimodal document understanding models, enhancing robustness and adaptability across diverse document-related VQA tasks.

\begin{table}[t]
\small
\begin{tabular}{lrrr}
\toprule
Dataset         & \#Images  & \#QA Pairs  & \#Tokens         \\ 
\midrule
Docmatix~\cite{laurenccon2024docmatix}                             & 2,444,750 & 9,500,000    & 390,000,000      \\

DocVQA~\cite{clark2017docqa}         & 10,189    & 39,463       & 337,829          \\

TextCaps~\cite{sidorov2020textcaps}  & 21,953    & 21,953       & 389,658          \\

TextVQA~\cite{singh2019textvqa}      & 21,953    & 34,602       & 181,918          \\

ST-VQA~\cite{biten2019stvqa}         & 17,247    & 23,121       & 127,846          \\

OCR-VQA~\cite{mishra2019ocrvqa}      & 165,746   & 801,579      & 6,073,824        \\

VisualMRC~\cite{tanaka2021visualmrc} & 3,027     & 11,988       & 168,828          \\

DUDE~\cite{van2023document}          & 147,597   & 23,716       & 11,341,228       \\ 
\hline
\rowcolor{gray!15}
\dataname (ours)                     & 3,103,494 & 758,000      & 5,200,000,000    \\ 
\bottomrule
\end{tabular}
\caption{
\textbf{Comparison with popular VQA datasets.}
}
\label{tab:dataset_comparison}

\vspace{-3mm}

\end{table}

\section{Enhanced Baseline for Document-Level Multimodal Understanding}

\subsection{Model Architecture}
Our model architecture leverages the widely-adopted ViT-MLP-LLM structure \cite{liu2023llava, liu2023improved, InternVL2}, consisting of a pre-trained Vision Transformer (ViT), a two-layer MLP projector, and a pre-trained Language Model (LLM). This combination provides a strong baseline for multimodal document analysis, effectively integrating visual and textual information within a unified framework.

\subsection{Optimizing Training Efficiency}

The training efficiency of MLLMs is hindered by two key challenges: (1) \emph{Inconsistent Sample Lengths.} Samples with different context lengths will result in excessive padding and lower training throughput; and (2) \emph{Limited GPU Memory.} 
As the model scale and context length increase, GPU memory consumption becomes increasingly unsustainable. To address these issues, we have implemented the following strategies:

(1) \emph{Multimodal Data Packing.} To balance the computational load between the vision model (ViT) and the language model (LLM) while minimizing resource waste caused by padding, we implement a multimodal data-packing strategy. The key idea is to concatenate multiple samples into long sequences to fully utilize the model's input capacity. 
Specifically, thresholds $T_{\rm img}$ and $T_{\rm tok}$ are set for the number of images and tokens, respectively. Samples are managed using a priority queue, sorted in descending order by the number of images and total tokens. A new sample \( s \) attempts to combine with the sample at the front of the priority queue. If the combination meets the thresholds (\ie, $T_{\rm img}$, $T_{\rm tok}$), the combined sample is pushed back into the priority queue. If \( s \) cannot match with any existing sample in the queue, it is directly added to the queue. When the image number and total token number of the front sample reach one of the thresholds or the number of samples exceeds the maximum limit \( M \), the front sample is dequeued, padded as needed, and sent for training. This strategy optimizes resource utilization and ensures balanced computational workloads. The detailed pseudo-code can be found in the supplementary materials.

(2) \emph{Ring Attention.} We implement the Ring Attention mechanism~~\cite{liu2023ring} to alleviate memory constraints associated with processing long sequences. By partitioning sequences into blocks and distributing computation across multiple devices, Ring Attention allows the model to accommodate larger contexts. This approach enables overlapping communication between key-value blocks and attention computations, thereby enhancing parallel processing efficiency. Consequently, Ring Attention improves the model's capacity to handle extended context lengths without exceeding memory limits.

(3) \emph{Liger Kernel.} To further improve memory and computational efficiency, we integrate the Liger Kernel~\cite{dai2024ligerkernel}, a specialized kernel library optimized for large-scale model training. The Liger Kernel enhances throughput and reduces memory consumption by employing techniques like kernel fusion, in-place operations, and input chunking. Leveraging the Liger Kernel thus enables higher training throughput and addresses memory limitations, allowing for efficient scaling of large multimodal models.

\begin{table*}[ht]
\centering
\footnotesize
\tabcolsep=3.3pt
\renewcommand{\arraystretch}{0.9}
    \begin{tabular}{lcccccccccccc}
    \toprule
    \multirow{2}{*}{Models}           
    & MP-Doc         & \multicolumn{2}{c}{MMLong-Doc}            & \multicolumn{5}{c}{DocGenome}
    & \multicolumn{4}{c}{MM-NIAH} \\
    \cmidrule(l){2-2}\cmidrule(l){3-4}\cmidrule(l){5-9}\cmidrule(l){10-13}
    & ANSL$\uparrow$ & Acc$\uparrow$  & F1$\uparrow$          & Class Acc$\uparrow$ & Title ED$\downarrow$ & Abstract ED$\downarrow$  & SP Acc$\uparrow$ & MP Acc$\uparrow$
    & Short    & Medium    & Long    & Overall \\
    \midrule
    \emph{Proprietary Models}\\
    Gemini-1.5-Pro~\cite{reid2024gemini1_5}  
    & --             & 28.2 & 20.6                            & --    & --   & --   & --    & -- 
    & 73.8 & 65.2 & 60.8 & 67.1 \\
    GPT-4o~\cite{gpt4v}       
    & --             & 42.8 & 44.9                            & 97.6 & 9.5 & 6.5 & 71.8 & 67.6 
    & -- & -- & -- & -- \\
    
    \midrule
    \emph{Open-Source Models}\\ 
    MiniMonkey-2B~\cite{huang2024minimonkey} 
    & 70.3           & 10.3 &  8.6                            & 57.4 & 16.5 & 55.0 & 40.3 & 28.9                         
    & 40.9 & 26.9 & 23.5 & 31.0 \\

    InternVL2-2B~\cite{chen2024far} 
    & 71.8           & 10.5 & 10.8                            & 60.8 & 18.4 & 54.3 & 39.4 & 28.9                       
    & 36.6 & 21.2 & 19.4 & 26.4 \\
    InternVL2-2B + RAG~\cite{wang2024needle} 
    & 72.6           & 17.2 & 16.7                            & 60.8 & 18.4 & 54.3 & 39.4 & 28.4 
    & 36.8 & 30.2 & 34.8 & 33.8 \\
    Llama3.2-3B-Instruct$^\dagger$~\cite{dubey2024llama3}
    & --             & 23.7 & 21.2                            & 85.3 & 194.7 & 51.0 & 40.2 & 34.9 
    & 15.5 &  2.2 &  0.5 &  6.6 \\
    \rowcolor{gray!15}
    \modelname-2B (ours)            
    & 76.2           & 21.8 & 16.0                            & 56.2 & 4.5  & 43.6 & 45.1 & 37.4  
    & 58.0 & 46.7 & 40.9 & 49.2 \\

    \midrule
    MiniCPM-V2.6-8B~\cite{yao2024minicpm_v} 
    & --             & 16.9 & 15.4                            & 92.8 & 10.2 & 32.6 & 60.0 & 54.2     
    & 49.0 & 15.3 & 0.0 & 23.4 \\

    LLaVA-OneVision-8B~\cite{li2024llavaonevision} 
    & --             & 10.8 &  9.6                            & 85.6 & 49.9 & 77.5 &  9.8 &  7.1 
    & 65.7 & 38.0 & 0.0 & 37.7 \\
    mPLUG-DocOwl2-8B~\cite{hu2024mplugdocowl2} 
    & 69.4           & 13.4 &  8.9                            & --   & --   & --   & --   & -- 
    & 17.9 &  0.1 & 0.0 &  6.6 \\
    M3DocRAG~\cite{cho2024m3docrag}           
    & 84.4           & 21.0 & 22.6                            & --   & --   & --   & --   & -- 
    & -- & -- & -- & -- \\

    VisRAG-8B~\cite{yu2024visrag}           
    & --             & 18.8 & 18.3                            & 92.8 & 10.2 & 32.6 & 60.0 & 50.7 
    & 47.1 & 29.2 & 29.5 & 35.8 \\
    
    InternLM2.5-7B-1M$^\dagger$~\cite{cai2024internlm2}   
    & --             & 28.7 & 25.6                            & 92.7 & 77.6 & 59.3 & 42.7 & 42.5 
    & 40.5 & 37.2 & 35.1 & 37.8 \\

    InternVL2-8B~\cite{chen2024far} 
    & 79.3           & 17.4 & 16.5                            & 90.6 &  8.2 & 39.6 & 56.0 & 46.1 
    & 56.4 & 37.3 & 32.4 & 42.9 \\
    InternVL2-8B + RAG~\cite{wang2024needle} 
    & 78.7           & 24.2 & 24.5                            & 90.6 &  8.2 & 39.6 & 56.0 & 46.0  
    & 55.7 & 43.4 & 45.2 & 48.4 \\

    InternVL2-26B~\cite{chen2024far} 
    & --             & 15.5 & 15.4                            & 87.5 & 16.9 & 23.3 & 49.7 & 42.7 
    & 65.0 & 48.7 & 41.9 & 52.8 \\

    \rowcolor{gray!15}
    \modelname-8B (ours)            
    & 81.3           & 28.8 & 23.0                            & 93.8 & 2.0  & 19.7 & 53.9 & 51.9 
    & 71.2 & 57.4 & 55.3 & 61.8 \\
    \bottomrule
    \end{tabular}
    \caption{
    \textbf{Evaluation on multi-page and interleaved VQA benchmarks.} 
    We report the metrics on MP-DocVQA~\cite{tito2023mpdocvqa} (MP-Doc), MMLongbench-Doc~\cite{ma2024mmlong} (MMLong-Doc), DocGenome~\cite{xia2024docgenome}, and MM-NIAH~\cite{wang2024needle}.
    Our model outperforms document-level MLLMs and multimodal RAG methods on multi-page, medium, and long-context QA. 
    The ``Short'', ``Medium'', and ``Long'' in MM-NIAH refer to input length in $\rm{[0,8k]}$, $\rm{(8k, 32k]}$, $\rm{( 32k, 64k]}$, respectively. ``$\dagger$'' denotes input documents are parsed by OCR models.}
    \label{tab:multi_page_qa}
    \vspace{-3mm}
\end{table*}

\section{Experiments}

\vspace{-0.5ex}

\subsection{Experimental Setup}
\vspace{-0.5ex}

\noindent\textbf{Training Details.}
Our model is available in two sizes: \modelname-2B and \modelname-8B, both of which are based on the InternVL2~\cite{InternVL2} and fine-tuned for one epoch using the data recipe that includes \dataname. 
The training uses a batch size of 128, the AdamW optimizer with a learning rate of 1e-5, weight decay of 0.01 for the 2B variant, and 0.05 for the 8B variant, along with a cosine learning rate schedule.
To speed up training, we apply multimodal data packing to reduce padding and a dynamic high-resolution strategy~\cite{chen2024far} to enhance OCR for document understanding. 
The maximum number of tiles for multimodal data is limited to 24, and the maximum sequence length is set to 32k tokens.

\noindent\textbf{Baselines.}
We compare our \modelname with a series of open-source document-level MLLMs~\cite{huang2024minimonkey,li2023monkey,liu2024textmonkey,hu2024mplugdocowl2,zhang2024internlm,yao2024minicpm_v,li2024llavaonevision} that supports multi-image input and proprietary MLLMs, including Gemini-1.5-pro~\cite{reid2024gemini1_5}, GPT-4o~\cite{gpt4o}. For comparison with the commonly used RAG method for handling long documents, we selected the latest multimodal RAG methods VisRAG~\cite{yu2024visrag}, InternVL + RAG~\cite{wang2024needle}, and M3DocRAG \cite{cho2024m3docrag}. To compare with more long-context large language models, we use InternVL2-8B as the OCR model to extract texts from the documents and images and feed the parsed documents to long-context LLMs~\cite{dubey2024llama3, cai2024internlm2}.

\begin{table}[t!]
\centering
\setlength{\tabcolsep}{7pt}
\footnotesize
    \begin{tabular}{lccc}
    \toprule
    Models & DocVQA & ChartQA & InfoVQA \\
    \midrule
    Gemini-1.5-Pro~\cite{reid2024gemini1_5}  & 93.1    & 87.2    & 81.0    \\
    GPT-4o~\cite{gpt4v}                      & 92.8    & 85.7    & --      \\
    \midrule
    MiniMonkey-2B~\cite{huang2024minimonkey} & 87.4    & 76.5    & 60.1    \\
    InternVL2-2B~\cite{chen2024far}          & 86.9    & 76.2    & 58.9    \\
    \rowcolor{gray!15}
    \modelname-2B (ours)                     & 87.3    & 76.4    & 58.5    \\
    \midrule
    Monkey-8B~\cite{li2023monkey}            & 66.5    & 65.1    & 36.1    \\
    TextMonkey-9B~\cite{liu2024textmonkey}   & 73.0    & 66.9    & 28.6    \\
    mPLUG-DocOwl2-8B~\cite{hu2024mplugdocowl2}& 80.7   & 70.0    & 46.4    \\
    IXC2.5-7B \cite{zhang2024internlm}       & 90.9    & 82.2    & 69.9    \\
    InternVL2-8B~\cite{chen2024far}          & 91.6    & 83.3    & 74.8    \\
    \rowcolor{gray!15}
    \modelname-8B (ours)                     & 92.0    & 83.3    & 73.3    \\
    \bottomrule
    \end{tabular}
    \vspace{-1mm}
    \caption{\textbf{Results on single-page VQA benchmarks.}
    Our \modelname models perform comparably to baselines~\cite{InternVL2}, demonstrating enhanced long-context modeling without loss on shorter tasks.
    }
    \label{tab:single_page_qa}
    \vspace{-3mm}
\end{table}

\vspace{-1ex}

\subsection{Multi-Page VQA}

\noindent\textbf{Benchmarks.} For the multi-page VQA task, we evaluate our model on three benchmarks:
(1) \textbf{MP-DocVQA} \cite{tito2023mpdocvqa}, 
which is designed to evaluate the ability to handle complex questions across multiple scanned document pages.
(2) \textbf{MMLongbench-Doc} \cite{ma2024mmlong}, a benchmark for evaluating the performance of MLLMs on multi-modal documents.
(3) \textbf{DocGenome} \cite{xia2024docgenome}, a large-scale benchmark for the evaluation of scientific document comprehension.

\noindent\textbf{Results.}
As illustrated in Table~\ref{tab:multi_page_qa}, our model achieves consistent improvements on multi-page QA benchmarks, outperforming previous document-level MLLMs.
Notably, our \modelname-8B surpasses Gemini-1.5-Pro \cite{reid2024gemini1_5} on MMLongBench-Doc, positioning it as the closest open-source model to GPT-4o. 
In comparison to RAG-based methods~\cite{yu2024visrag, cho2024m3docrag, wang2024mmniah}, our model demonstrates advantages in multi-page scenarios.
For example, in the Multi-Page QA of DocGenome benchmark, the RAG method shows a performance decline due to the disruption of document continuity while our \modelname exhibits a significantly stable improvement compared to the baseline, with \modelname-8B showing an increase of 12.6\% over InternVL2-8B.

\subsection{Interleaved Long-Context QA}

\noindent\textbf{Benchmarks.}
For the interleaved long-context QA task, we evaluate our models on MM-NIAH~\cite{wang2024mmniah}, a benchmark designed for long multimodal document comprehension.

\noindent\textbf{Results.}
The right side of Table~\ref{tab:multi_page_qa} presents the results of MM-NIAH across context lengths ranging from 1K to 64K. We categorize the context lengths into "Short," "Medium," and "Long" based on the context window of InternVL2 (8K) and \modelname (32K). Our \modelname demonstrates exceptional performance in both medium- and long-context scenarios, while maintaining high accuracy in short-context situations. Notably, for QA tasks with context lengths in the range of $(32K, 64K]$, \modelname-2B outperforms InternVL2-2B by 110\%, and \modelname-8B surpasses InternVL2-8B by 70\%. Furthermore, our model performs comparably to the state-of-the-art multimodal long-context model Gemini-1.5-Pro in contexts longer than 8K, establishing a new state-of-the-art performance among open-source long-context MLLMs.

\subsection{Single-Page VQA}

\noindent\textbf{Benchmarks.}
For single-page VQA tasks, we evaluate our model on three benchmarks:
(1) DocVQA \cite{mathew2021docvqa}, a benchmark for the evaluation of extracting key information from an image of the given document.
(2) ChartQA \cite{masry2022chartqa}, a benchmark for evaluating the reasoning abilities for chart images.
(3) InfoVQA \cite{mathew2022infographicvqa}, a benchmark for infographic image comprehension.

\noindent\textbf{Results.}
As shown in Table~\ref{tab:single_page_qa}, our model achieves 
comparable performance to baseline models. 
Across the three benchmarks, \modelname-2B and InternVL2-2B exhibit comparable results, while \modelname-8B outperforms InternVL2-8B by 0.4 points in DocVQA. These results demonstrate that \dataname effectively enhances the model's long-context modeling capabilities without compromising its performance on shorter documents.

\subsection{Ablation Study}

\noindent\textbf{Effect of \dataname.}
\begin{table}[t]
\centering
\setlength{\tabcolsep}{8pt}
\footnotesize
    \begin{tabular}{lcc}
    \toprule
    Models                                                & Acc    & F1    \\
    \midrule
    Baseline (InternVL2-2B~\cite{InternVL2})              & 10.5   & 10.8  \\
    -- Variant1: SFT using data recipe w/o \dataname      & 18.4   & 9.4   \\
    -- Variant2: Variant1 + our Sci-Hub data              & 18.5   & 15.2  \\
    -- Variant3: Variant2 + our Arxiv data                & 20.5   & 15.5  \\
    \rowcolor{gray!15}
    \modelname-2B: Variant3 + our OpenReview data         & 21.8   & 16.0  \\
    \bottomrule
    \end{tabular}
    \caption{\textbf{Ablation study on the data recipe.}
    We evaluate the effectiveness of the training data from different sources on MMLongBench-Doc. Our \dataname can consistently enhance the ability of the model to understand multi-page documents.}
    \label{tab:data_effect}
\end{table}
We conducted ablation studies on MMLongBench-Doc~\cite{ma2024mmlong} to analyze the impact of our \dataname. We divided \dataname into 3 parts according to the source of the data: (1) Sci-Hub data; (2) Arxiv data; and (3) OpenReview data. We demonstrate the effects of incorporating each part of the data into the SFT process, reported in Table~\ref{tab:data_effect}. 
We observed that with the inclusion of different parts of \dataname, the model's performance improves continuously. Utilizing only open-source data results in an inferior F1 score.

\begin{table}[t]
\centering
\setlength{\tabcolsep}{10pt}
\footnotesize
    \begin{tabular}{lrrr}
    \toprule
    Models                                    & Latency               & Acc   & F1   \\
    \midrule
    MiniCPM-V2.6~\cite{yao2024minicpm_v}      &  225.4ms              & 16.9  & 15.4 \\
    VisRAG-12B~\cite{yu2024visrag}            &  288.3ms              & 18.8  & 18.3 \\
    \midrule
    InternVL2-2B~\cite{chen2024far}           &   35.9ms              & 10.5  &  10.8 \\
    InternVL2-2B + RAG~\cite{wang2024needle}&   82.9ms              & 17.2  &  16.7 \\
    \rowcolor{gray!15}
    \modelname-2B (ours)                      &   35.9ms              & 21.8  &  16.0 \\
    \midrule
    InternVL2-8B~\cite{chen2024far}           &   81.0ms              & 17.4  &  16.5 \\
    InternVL2-8B + RAG~\cite{wang2024needle}&  113.4ms              & 24.2  &  24.5 \\
    \rowcolor{gray!15}
    \modelname-8B (ours)                      &   81.0ms              & 28.8  &  23.0 \\
    \bottomrule
    \end{tabular}
    \caption{\textbf{Latency analysis.} We evaluate the average token output latency of model outputs on MMLongBench-Doc~\cite{ma2024mmlong}. 
    RAG-based methods~\cite{wang2024needle, yu2024visrag} exhibit slower processing speeds due to their two-stage inference process, making them less efficient than document MLLMs for handling multimodal long documents.
    }
    \label{tab:latency}
\end{table}
\noindent\textbf{Latency Analysis.}
To compare the latency in inference between RAG methods and our \modelname, we conducted a latency analysis on MMLongBench-Doc~\cite{ma2024mmlong}, as reported in Table~\ref{tab:latency}. While RAG reduces the document length input to the MLLM, its own time cost remains non-negligible. For instance, InternVL2-2B + RAG is 130\% slower than InternVL2-2B, and VisRAG is 28\% slower than MiniCPM-V2.6. Our \modelname does not require additional processes and therefore has the same inference time as baseline models, making it more suitable for analyzing long documents.

\section{Conclusions}

This work introduced a diverse document-level question-answering dataset that covers complex structures and cross-page dependencies, providing a robust foundation for training and evaluating document understanding models. We also proposed a retrieval-free long-document understanding model that effectively integrates multi-page information, reducing reliance on external retrieval systems. Experimental results show that our model achieves state-of-the-art performance across several document-level QA benchmarks, underscoring its strength in multi-page integration and complex reasoning. Future work will focus on improving computational efficiency, extending the model to larger multimodal tasks, and adapting it to broader applications for enhanced practicality and generalization.

\section*{Acknowledgments}
This project was supported by the National Key R\&D Program of China (No. 2022ZD0161300, 2022ZD0160101), the National Natural Science Foundation of China (No. 62376134, 62372223). Zhe Chen is supported by the Youth PhD Student Research Project under the National Natural Science Foundation (No. 623B2050).

{
    \small
    \bibliographystyle{ieeenat_fullname}
    \bibliography{main}

\begin{thebibliography}{99}
\providecommand{\natexlab}[1]{#1}
\providecommand{\url}[1]{\texttt{#1}}
\expandafter\ifx\csname urlstyle\endcsname\relax
  \providecommand{\doi}[1]{doi: #1}\else
  \providecommand{\doi}{doi: \begingroup \urlstyle{rm}\Url}\fi

\bibitem[Alayrac et~al.(2022)Alayrac, Donahue, Luc, Miech, Barr, Hasson, Lenc, Mensch, Millican, Reynolds, et~al.]{alayrac2022flamingo}
Jean-Baptiste Alayrac, Jeff Donahue, Pauline Luc, Antoine Miech, Iain Barr, Yana Hasson, Karel Lenc, Arthur Mensch, Katherine Millican, Malcolm Reynolds, et~al.
\newblock Flamingo: a visual language model for few-shot learning.
\newblock \emph{NeurIPS}, 35:\penalty0 23716--23736, 2022.

\bibitem[An et~al.(2023)An, Gong, Zhong, Zhao, Li, Zhang, Kong, and Qiu]{an2023leval}
Chenxin An, Shansan Gong, Ming Zhong, Xingjian Zhao, Mukai Li, Jun Zhang, Lingpeng Kong, and Xipeng Qiu.
\newblock L-eval: Instituting standardized evaluation for long context language models.
\newblock \emph{arXiv preprint arXiv:2307.11088}, 2023.

\bibitem[Appalaraju et~al.(2024)Appalaraju, Tang, Dong, Sankaran, Zhou, and Manmatha]{appalaraju2024docformerv2}
Srikar Appalaraju, Peng Tang, Qi Dong, Nishant Sankaran, Yichu Zhou, and R Manmatha.
\newblock Docformerv2: Local features for document understanding.
\newblock In \emph{Proceedings of the AAAI Conference on Artificial Intelligence}, pages 709--718, 2024.

\bibitem[Bai et~al.(2022)Bai, Liu, Meng, Li, Liu, Xie, Zheng, Wang, Hou, Wei, et~al.]{bai2022wukong}
Haoli Bai, Zhiguang Liu, Xiaojun Meng, Wentao Li, Shuang Liu, Nian Xie, Rongfu Zheng, Liangwei Wang, Lu Hou, Jiansheng Wei, et~al.
\newblock Wukong-reader: Multi-modal pre-training for fine-grained visual document understanding.
\newblock \emph{arXiv preprint arXiv:2212.09621}, 2022.

\bibitem[Bai et~al.(2023{\natexlab{a}})Bai, Bai, Chu, Cui, Dang, Deng, Fan, Ge, Han, Huang, Hui, Ji, Li, Lin, Lin, Liu, Liu, Lu, Lu, Ma, Men, Ren, Ren, Tan, Tan, Tu, Wang, Wang, Wang, Wu, Xu, Xu, Yang, Yang, Yang, Yang, Yao, Yu, Yuan, Yuan, Zhang, Zhang, Zhang, Zhang, Zhou, Zhou, Zhou, and Zhu]{qwen}
Jinze Bai, Shuai Bai, Yunfei Chu, Zeyu Cui, Kai Dang, Xiaodong Deng, Yang Fan, Wenbin Ge, Yu Han, Fei Huang, Binyuan Hui, Luo Ji, Mei Li, Junyang Lin, Runji Lin, Dayiheng Liu, Gao Liu, Chengqiang Lu, Keming Lu, Jianxin Ma, Rui Men, Xingzhang Ren, Xuancheng Ren, Chuanqi Tan, Sinan Tan, Jianhong Tu, Peng Wang, Shijie Wang, Wei Wang, Shengguang Wu, Benfeng Xu, Jin Xu, An Yang, Hao Yang, Jian Yang, Shusheng Yang, Yang Yao, Bowen Yu, Hongyi Yuan, Zheng Yuan, Jianwei Zhang, Xingxuan Zhang, Yichang Zhang, Zhenru Zhang, Chang Zhou, Jingren Zhou, Xiaohuan Zhou, and Tianhang Zhu.
\newblock Qwen technical report.
\newblock \emph{arXiv preprint arXiv:2309.16609}, 2023{\natexlab{a}}.

\bibitem[Bai et~al.(2023{\natexlab{b}})Bai, Bai, Yang, Wang, Tan, Wang, Lin, Zhou, and Zhou]{bai2023qwenvl}
Jinze Bai, Shuai Bai, Shusheng Yang, Shijie Wang, Sinan Tan, Peng Wang, Junyang Lin, Chang Zhou, and Jingren Zhou.
\newblock Qwen-vl: A frontier large vision-language model with versatile abilities.
\newblock \emph{arXiv preprint arXiv:2308.12966}, 2023{\natexlab{b}}.

\bibitem[Bai et~al.(2024)Bai, Lv, Zhang, He, Qi, Hou, Tang, Dong, and Li]{bai2024longalign}
Yushi Bai, Xin Lv, Jiajie Zhang, Yuze He, Ji Qi, Lei Hou, Jie Tang, Yuxiao Dong, and Juanzi Li.
\newblock Longalign: A recipe for long context alignment of large language models.
\newblock \emph{arXiv preprint arXiv:2401.18058}, 2024.

\bibitem[Biten et~al.(2019)Biten, Tito, Mafla, Gomez, Rusinol, Valveny, Jawahar, and Karatzas]{biten2019stvqa}
Ali~Furkan Biten, Ruben Tito, Andres Mafla, Lluis Gomez, Mar{\c{c}}al Rusinol, Ernest Valveny, CV Jawahar, and Dimosthenis Karatzas.
\newblock Scene text visual question answering.
\newblock In \emph{ICCV}, pages 4291--4301, 2019.

\bibitem[Cai et~al.(2024)Cai, Cao, Chen, Chen, Chen, Chen, Chen, Chen, Chen, Chu, et~al.]{cai2024internlm2}
Zheng Cai, Maosong Cao, Haojiong Chen, Kai Chen, Keyu Chen, Xin Chen, Xun Chen, Zehui Chen, Zhi Chen, Pei Chu, et~al.
\newblock Internlm2 technical report.
\newblock \emph{arXiv preprint arXiv:2403.17297}, 2024.

\bibitem[Chen et~al.(2015)Chen, Fang, Lin, Vedantam, Gupta, Doll{\'a}r, and Zitnick]{chen2015cococaption}
Xinlei Chen, Hao Fang, Tsung-Yi Lin, Ramakrishna Vedantam, Saurabh Gupta, Piotr Doll{\'a}r, and C~Lawrence Zitnick.
\newblock Microsoft coco captions: Data collection and evaluation server.
\newblock \emph{arXiv preprint arXiv:1504.00325}, 2015.

\bibitem[Chen et~al.(2023{\natexlab{a}})Chen, Qian, Tang, Lai, Liu, Han, and Jia]{chen2023longlora}
Yukang Chen, Shengju Qian, Haotian Tang, Xin Lai, Zhijian Liu, Song Han, and Jiaya Jia.
\newblock Longlora: Efficient fine-tuning of long-context large language models.
\newblock \emph{arXiv preprint arXiv:2309.12307}, 2023{\natexlab{a}}.

\bibitem[Chen et~al.(2023{\natexlab{b}})Chen, Wu, Wang, Su, Chen, Xing, Muyan, Zhang, Zhu, Lu, et~al.]{chen2023internvl}
Zhe Chen, Jiannan Wu, Wenhai Wang, Weijie Su, Guo Chen, Sen Xing, Zhong Muyan, Qinglong Zhang, Xizhou Zhu, Lewei Lu, et~al.
\newblock Internvl: Scaling up vision foundation models and aligning for generic visual-linguistic tasks.
\newblock \emph{arXiv preprint arXiv:2312.14238}, 2023{\natexlab{b}}.

\bibitem[Chen et~al.(2024)Chen, Wang, Tian, Ye, Gao, Cui, Tong, Hu, Luo, Ma, et~al.]{chen2024far}
Zhe Chen, Weiyun Wang, Hao Tian, Shenglong Ye, Zhangwei Gao, Erfei Cui, Wenwen Tong, Kongzhi Hu, Jiapeng Luo, Zheng Ma, et~al.
\newblock How far are we to gpt-4v? closing the gap to commercial multimodal models with open-source suites.
\newblock \emph{arXiv preprint arXiv:2404.16821}, 2024.

\bibitem[Cho et~al.(2024)Cho, Mahata, Irsoy, He, and Bansal]{cho2024m3docrag}
Jaemin Cho, Debanjan Mahata, Ozan Irsoy, Yujie He, and Mohit Bansal.
\newblock M3docrag: Multi-modal retrieval is what you need for multi-page multi-document understanding.
\newblock \emph{arXiv preprint arXiv:2411.04952}, 2024.

\bibitem[Clark and Gardner(2018)]{clark2017docqa}
Christopher Clark and Matt Gardner.
\newblock Simple and effective multi-paragraph reading comprehension.
\newblock In \emph{ACL}, pages 845--855, 2018.

\bibitem[Dai et~al.(2024)Dai, Kothapalli, Song, Tang, Zhu, Shimizu, Sahni, Ning, Chen, et~al.]{dai2024ligerkernel}
Yun Dai, Vignesh Kothapalli, Qingquan Song, Shao Tang, Siyu Zhu, Steven Shimizu, Shivam Sahni, Haowen Ning, Yanning Chen, et~al.
\newblock Liger kernel: Efficient triton kernels for llm training.
\newblock \emph{arXiv preprint arXiv:2410.10989}, 2024.

\bibitem[Dao(2023)]{dao2023flashattention}
Tri Dao.
\newblock Flashattention-2: Faster attention with better parallelism and work partitioning.
\newblock \emph{arXiv preprint arXiv:2307.08691}, 2023.

\bibitem[Dao et~al.(2022)Dao, Fu, Ermon, Rudra, and R{\'e}]{dao2022flashattention}
Tri Dao, Dan Fu, Stefano Ermon, Atri Rudra, and Christopher R{\'e}.
\newblock Flashattention: Fast and memory-efficient exact attention with io-awareness.
\newblock \emph{NeurIPS}, 35:\penalty0 16344--16359, 2022.

\bibitem[Ding et~al.(2023)Ding, Ma, Dong, Zhang, Huang, Wang, Zheng, and Wei]{ding2023longnet}
Jiayu Ding, Shuming Ma, Li Dong, Xingxing Zhang, Shaohan Huang, Wenhui Wang, Nanning Zheng, and Furu Wei.
\newblock Longnet: Scaling transformers to 1,000,000,000 tokens.
\newblock \emph{arXiv preprint arXiv:2307.02486}, 2023.

\bibitem[Dong et~al.(2024)Dong, Han, Peng, Qi, Ge, Yang, Zhao, Sun, Zhou, Wei, et~al.]{dong2023dreamllm}
Runpei Dong, Chunrui Han, Yuang Peng, Zekun Qi, Zheng Ge, Jinrong Yang, Liang Zhao, Jianjian Sun, Hongyu Zhou, Haoran Wei, et~al.
\newblock Dreamllm: Synergistic multimodal comprehension and creation.
\newblock In \emph{ICLR}, 2024.

\bibitem[Dubey et~al.(2024)Dubey, Jauhri, Pandey, Kadian, Al-Dahle, Letman, Mathur, Schelten, Yang, Fan, et~al.]{dubey2024llama3}
Abhimanyu Dubey, Abhinav Jauhri, Abhinav Pandey, Abhishek Kadian, Ahmad Al-Dahle, Aiesha Letman, Akhil Mathur, Alan Schelten, Amy Yang, Angela Fan, et~al.
\newblock The llama 3 herd of models.
\newblock \emph{arXiv preprint arXiv:2407.21783}, 2024.

\bibitem[Faysse et~al.(2024)Faysse, Sibille, Wu, Viaud, Hudelot, and Colombo]{faysse2024colpali}
Manuel Faysse, Hugues Sibille, Tony Wu, Gautier Viaud, C{\'e}line Hudelot, and Pierre Colombo.
\newblock Colpali: Efficient document retrieval with vision language models.
\newblock \emph{arXiv preprint arXiv:2407.01449}, 2024.

\bibitem[Feng et~al.(2023)Feng, Liu, Liu, Zhou, Li, and Huang]{feng2023docpedia}
Hao Feng, Qi Liu, Hao Liu, Wengang Zhou, Houqiang Li, and Can Huang.
\newblock Docpedia: Unleashing the power of large multimodal model in the frequency domain for versatile document understanding.
\newblock \emph{arXiv preprint arXiv:2311.11810}, 2023.

\bibitem[Fu et~al.(2023)Fu, Chen, Shen, Qin, Zhang, Lin, Qiu, Lin, Yang, Zheng, et~al.]{fu2023mme}
Chaoyou Fu, Peixian Chen, Yunhang Shen, Yulei Qin, Mengdan Zhang, Xu Lin, Zhenyu Qiu, Wei Lin, Jinrui Yang, Xiawu Zheng, et~al.
\newblock Mme: A comprehensive evaluation benchmark for multimodal large language models.
\newblock \emph{arXiv preprint arXiv:2306.13394}, 2023.

\bibitem[Gao et~al.(2024)Gao, Chen, Cui, Ren, Wang, Zhu, Tian, Ye, He, Zhu, et~al.]{gao2024mini_internvl}
Zhangwei Gao, Zhe Chen, Erfei Cui, Yiming Ren, Weiyun Wang, Jinguo Zhu, Hao Tian, Shenglong Ye, Junjun He, Xizhou Zhu, et~al.
\newblock Mini-internvl: A flexible-transfer pocket multimodal model with 5\% parameters and 90\% performance.
\newblock \emph{arXiv preprint arXiv:2410.16261}, 2024.

\bibitem[Han et~al.(2023)Han, Wang, Xiong, Chen, Ji, and Wang]{han2023lm}
Chi Han, Qifan Wang, Wenhan Xiong, Yu Chen, Heng Ji, and Sinong Wang.
\newblock Lm-infinite: Simple on-the-fly length generalization for large language models.
\newblock \emph{arXiv preprint arXiv:2308.16137}, 2023.

\bibitem[Hu et~al.(2024)Hu, Xu, Zhang, Ye, Yan, Zhang, Jin, Huang, and Zhou]{hu2024mplugdocowl2}
Anwen Hu, Haiyang Xu, Liang Zhang, Jiabo Ye, Ming Yan, Ji Zhang, Qin Jin, Fei Huang, and Jingren Zhou.
\newblock mplug-docowl2: High-resolution compressing for ocr-free multi-page document understanding.
\newblock \emph{arXiv preprint arXiv:2409.03420}, 2024.

\bibitem[Huang et~al.(2024)Huang, Liu, Liang, Jin, and Bai]{huang2024minimonkey}
Mingxin Huang, Yuliang Liu, Dingkang Liang, Lianwen Jin, and Xiang Bai.
\newblock Mini-monkey: Alleviate the sawtooth effect by multi-scale adaptive cropping.
\newblock \emph{arXiv preprint arXiv:2408.02034}, 2024.

\bibitem[Ilharco et~al.(2021)Ilharco, Wortsman, Wightman, Gordon, Carlini, Taori, Dave, Shankar, Namkoong, Miller, Hajishirzi, Farhadi, and Schmidt]{openclip}
Gabriel Ilharco, Mitchell Wortsman, Ross Wightman, Cade Gordon, Nicholas Carlini, Rohan Taori, Achal Dave, Vaishaal Shankar, Hongseok Namkoong, John Miller, Hannaneh Hajishirzi, Ali Farhadi, and Ludwig Schmidt.
\newblock Openclip.
\newblock Zenodo. Version 0.1. \url{https://doi.org/10.5281/zenodo.5143773}, 2021.
\newblock DOI: 10.5281/zenodo.5143773.

\bibitem[Jiang et~al.(2024)Jiang, Yan, Ji, Fu, Sun, Xiong, Fan, and Khan]{jiang2024effectiveness}
Yao Jiang, Xinyu Yan, Ge-Peng Ji, Keren Fu, Meijun Sun, Huan Xiong, Deng-Ping Fan, and Fahad~Shahbaz Khan.
\newblock Effectiveness assessment of recent large vision-language models.
\newblock \emph{Visual Intelligence}, 2\penalty0 (1):\penalty0 17, 2024.

\bibitem[Kamradt(2024)]{LLMTest_NeedleInAHaystack}
Greg Kamradt.
\newblock Llmtest\_needleinahaystack.
\newblock \url{https://github.com/gkamradt/LLMTest_NeedleInAHaystack}, 2024.
\newblock Accessed: 2024-11-11.

\bibitem[Kim et~al.(2022)Kim, Hong, Yim, Nam, Park, Yim, Hwang, Yun, Han, and Park]{kim2022ocr}
Geewook Kim, Teakgyu Hong, Moonbin Yim, JeongYeon Nam, Jinyoung Park, Jinyeong Yim, Wonseok Hwang, Sangdoo Yun, Dongyoon Han, and Seunghyun Park.
\newblock Ocr-free document understanding transformer.
\newblock In \emph{European Conference on Computer Vision}, pages 498--517. Springer, 2022.

\bibitem[Lauren{\c{c}}on et~al.(2024{\natexlab{a}})Lauren{\c{c}}on, Marafioti, Sanh, and Tronchon]{laurenccon2024docmatix}
Hugo Lauren{\c{c}}on, Andr{\'e}s Marafioti, Victor Sanh, and L{\'e}o Tronchon.
\newblock Building and better understanding vision-language models: insights and future directions.
\newblock \emph{arXiv preprint arXiv:2408.12637}, 2024{\natexlab{a}}.

\bibitem[Lauren{\c{c}}on et~al.(2024{\natexlab{b}})Lauren{\c{c}}on, Saulnier, Tronchon, Bekman, Singh, Lozhkov, Wang, Karamcheti, Rush, Kiela, et~al.]{laurenccon2024obelics}
Hugo Lauren{\c{c}}on, Lucile Saulnier, L{\'e}o Tronchon, Stas Bekman, Amanpreet Singh, Anton Lozhkov, Thomas Wang, Siddharth Karamcheti, Alexander Rush, Douwe Kiela, et~al.
\newblock Obelics: An open web-scale filtered dataset of interleaved image-text documents.
\newblock \emph{NIPS}, 36, 2024{\natexlab{b}}.

\bibitem[Li et~al.(2023{\natexlab{a}})Li, Wang, Wang, Ge, Ge, and Shan]{li2023seed}
Bohao Li, Rui Wang, Guangzhi Wang, Yuying Ge, Yixiao Ge, and Ying Shan.
\newblock Seed-bench: Benchmarking multimodal llms with generative comprehension.
\newblock \emph{arXiv preprint arXiv:2307.16125}, 2023{\natexlab{a}}.

\bibitem[Li et~al.(2024{\natexlab{a}})Li, Zhang, Guo, Zhang, Li, Zhang, Zhang, Li, Liu, and Li]{li2024llavaonevision}
Bo Li, Yuanhan Zhang, Dong Guo, Renrui Zhang, Feng Li, Hao Zhang, Kaichen Zhang, Yanwei Li, Ziwei Liu, and Chunyuan Li.
\newblock Llava-onevision: Easy visual task transfer.
\newblock \emph{arXiv preprint arXiv:2408.03326}, 2024{\natexlab{a}}.

\bibitem[Li et~al.(2022)Li, Li, Xiong, and Hoi]{li2022blip}
Junnan Li, Dongxu Li, Caiming Xiong, and Steven Hoi.
\newblock Blip: Bootstrapping language-image pre-training for unified vision-language understanding and generation.
\newblock In \emph{ICML}, pages 12888--12900, 2022.

\bibitem[Li et~al.(2023{\natexlab{b}})Li, Li, Savarese, and Hoi]{li2023blip2}
Junnan Li, Dongxu Li, Silvio Savarese, and Steven Hoi.
\newblock Blip-2: Bootstrapping language-image pre-training with frozen image encoders and large language models.
\newblock In \emph{ICML}, pages 19730--19742. PMLR, 2023{\natexlab{b}}.

\bibitem[Li et~al.(2024{\natexlab{b}})Li, Chen, Wang, Wang, Ye, Jin, Chen, He, Gao, Cui, et~al.]{li2024omnicorpus}
Qingyun Li, Zhe Chen, Weiyun Wang, Wenhai Wang, Shenglong Ye, Zhenjiang Jin, Guanzhou Chen, Yinan He, Zhangwei Gao, Erfei Cui, et~al.
\newblock Omnicorpus: An unified multimodal corpus of 10 billion-level images interleaved with text.
\newblock \emph{arXiv preprint arXiv:2406.08418}, 2024{\natexlab{b}}.

\bibitem[Li et~al.(2023{\natexlab{c}})Li, Yang, Liu, Ma, Zhang, Yang, Sun, Liu, and Bai]{li2023monkey}
Zhang Li, Biao Yang, Qiang Liu, Zhiyin Ma, Shuo Zhang, Jingxu Yang, Yabo Sun, Yuliang Liu, and Xiang Bai.
\newblock Monkey: Image resolution and text label are important things for large multi-modal models.
\newblock \emph{arXiv preprint arXiv:2311.06607}, 2023{\natexlab{c}}.

\bibitem[Liu et~al.(2023{\natexlab{a}})Liu, Li, Li, and Lee]{liu2023improved}
Haotian Liu, Chunyuan Li, Yuheng Li, and Yong~Jae Lee.
\newblock Improved baselines with visual instruction tuning.
\newblock \emph{arXiv preprint arXiv:2310.03744}, 2023{\natexlab{a}}.

\bibitem[Liu et~al.(2023{\natexlab{b}})Liu, Li, Wu, and Lee]{liu2023llava}
Haotian Liu, Chunyuan Li, Qingyang Wu, and Yong~Jae Lee.
\newblock Visual instruction tuning.
\newblock \emph{NeurIPS}, 36, 2023{\natexlab{b}}.

\bibitem[Liu et~al.(2023{\natexlab{c}})Liu, Zaharia, and Abbeel]{liu2023ring}
Hao Liu, Matei Zaharia, and Pieter Abbeel.
\newblock Ring attention with blockwise transformers for near-infinite context.
\newblock \emph{arXiv preprint arXiv:2310.01889}, 2023{\natexlab{c}}.

\bibitem[Liu et~al.(2023{\natexlab{d}})Liu, Duan, Zhang, Li, Zhang, Zhao, Yuan, Wang, He, Liu, et~al.]{liu2023mmbench}
Yuan Liu, Haodong Duan, Yuanhan Zhang, Bo Li, Songyang Zhang, Wangbo Zhao, Yike Yuan, Jiaqi Wang, Conghui He, Ziwei Liu, et~al.
\newblock Mmbench: Is your multi-modal model an all-around player?
\newblock \emph{arXiv preprint arXiv:2307.06281}, 2023{\natexlab{d}}.

\bibitem[Liu et~al.(2023{\natexlab{e}})Liu, Li, Li, Yu, Huang, Peng, Liu, Chen, Li, Jin, et~al.]{liu2023ocrbench}
Yuliang Liu, Zhang Li, Hongliang Li, Wenwen Yu, Mingxin Huang, Dezhi Peng, Mingyu Liu, Mingrui Chen, Chunyuan Li, Lianwen Jin, et~al.
\newblock On the hidden mystery of ocr in large multimodal models.
\newblock \emph{arXiv preprint arXiv:2305.07895}, 2023{\natexlab{e}}.

\bibitem[Liu et~al.(2024{\natexlab{a}})Liu, Yang, Liu, Li, Ma, Zhang, and Bai]{liu2024textmonkey}
Yuliang Liu, Biao Yang, Qiang Liu, Zhang Li, Zhiyin Ma, Shuo Zhang, and Xiang Bai.
\newblock Textmonkey: An ocr-free large multimodal model for understanding document.
\newblock \emph{arXiv preprint arXiv:2403.04473}, 2024{\natexlab{a}}.

\bibitem[Liu et~al.(2023{\natexlab{f}})Liu, He, Wang, Wang, Wang, Chen, Zhang, Lai, Yang, Li, Yu, et~al.]{2023interngpt}
Zhaoyang Liu, Yinan He, Wenhai Wang, Weiyun Wang, Yi Wang, Shoufa Chen, Qinglong Zhang, Zeqiang Lai, Yang Yang, Qingyun Li, Jiashuo Yu, et~al.
\newblock Interngpt: Solving vision-centric tasks by interacting with chatgpt beyond language.
\newblock \emph{arXiv preprint arXiv:2305.05662}, 2023{\natexlab{f}}.

\bibitem[Liu et~al.(2024{\natexlab{b}})Liu, Chu, Zang, Wei, Dong, Zhang, Liang, Xiong, Qiao, Lin, et~al.]{liu2024mmdu}
Ziyu Liu, Tao Chu, Yuhang Zang, Xilin Wei, Xiaoyi Dong, Pan Zhang, Zijian Liang, Yuanjun Xiong, Yu Qiao, Dahua Lin, et~al.
\newblock Mmdu: A multi-turn multi-image dialog understanding benchmark and instruction-tuning dataset for lvlms.
\newblock \emph{arXiv preprint arXiv:2406.11833}, 2024{\natexlab{b}}.

\bibitem[Luo et~al.(2024)Luo, Shen, Zhu, Zheng, Yu, and Yao]{luo2024layoutllm}
Chuwei Luo, Yufan Shen, Zhaoqing Zhu, Qi Zheng, Zhi Yu, and Cong Yao.
\newblock Layoutllm: Layout instruction tuning with large language models for document understanding.
\newblock In \emph{Proceedings of the IEEE/CVF Conference on Computer Vision and Pattern Recognition}, pages 15630--15640, 2024.

\bibitem[Ma et~al.(2024{\natexlab{a}})Ma, Lin, Li, Chen, and Lin]{ma2024unifying}
Xueguang Ma, Sheng-Chieh Lin, Minghan Li, Wenhu Chen, and Jimmy Lin.
\newblock Unifying multimodal retrieval via document screenshot embedding.
\newblock \emph{arXiv preprint arXiv:2406.11251}, 2024{\natexlab{a}}.

\bibitem[Ma et~al.(2024{\natexlab{b}})Ma, Zang, Chen, Chen, Jiao, Li, Lu, Liu, Ma, Dong, Zhang, Pan, Jiang, Wang, Cao, and Sun]{ma2024mmlong}
Yubo Ma, Yuhang Zang, Liangyu Chen, Meiqi Chen, Yizhu Jiao, Xinze Li, Xinyuan Lu, Ziyu Liu, Yan Ma, Xiaoyi Dong, Pan Zhang, Liangming Pan, Yu-Gang Jiang, Jiaqi Wang, Yixin Cao, and Aixin Sun.
\newblock Mmlongbench-doc: Benchmarking long-context document understanding with visualizations, 2024{\natexlab{b}}.

\bibitem[Masry et~al.(2022)Masry, Do, Tan, Joty, and Hoque]{masry2022chartqa}
Ahmed Masry, Xuan~Long Do, Jia~Qing Tan, Shafiq Joty, and Enamul Hoque.
\newblock Chartqa: A benchmark for question answering about charts with visual and logical reasoning.
\newblock In \emph{ACL}, pages 2263--2279, 2022.

\bibitem[Mathew et~al.(2021)Mathew, Karatzas, and Jawahar]{mathew2021docvqa}
Minesh Mathew, Dimosthenis Karatzas, and CV Jawahar.
\newblock Docvqa: A dataset for vqa on document images.
\newblock In \emph{WACV}, pages 2200--2209, 2021.

\bibitem[Mathew et~al.(2022)Mathew, Bagal, Tito, Karatzas, Valveny, and Jawahar]{mathew2022infographicvqa}
Minesh Mathew, Viraj Bagal, Rub{\`e}n Tito, Dimosthenis Karatzas, Ernest Valveny, and CV Jawahar.
\newblock Infographicvqa.
\newblock In \emph{WACV}, pages 1697--1706, 2022.

\bibitem[Mishra et~al.(2019)Mishra, Shekhar, Singh, and Chakraborty]{mishra2019ocrvqa}
Anand Mishra, Shashank Shekhar, Ajeet~Kumar Singh, and Anirban Chakraborty.
\newblock Ocr-vqa: Visual question answering by reading text in images.
\newblock In \emph{ICDAR}, pages 947--952, 2019.

\bibitem[OpenAI(2023)]{gpt4v}
OpenAI.
\newblock Gpt-4v(ision) system card.
\newblock \url{https://cdn.openai.com/papers/GPTV_System_Card.pdf}, 2023.

\bibitem[OpenAI(2024)]{gpt4o}
OpenAI.
\newblock Gpt-4o system card.
\newblock \url{https://openai.com/index/gpt-4o-system-card/}, 2024.

\bibitem[Press et~al.(2021)Press, Smith, and Lewis]{press2021train}
Ofir Press, Noah~A Smith, and Mike Lewis.
\newblock Train short, test long: Attention with linear biases enables input length extrapolation.
\newblock \emph{arXiv preprint arXiv:2108.12409}, 2021.

\bibitem[Rebuffi et~al.(2017)Rebuffi, Bilen, and Vedaldi]{rebuffi2017learning}
Sylvestre-Alvise Rebuffi, Hakan Bilen, and Andrea Vedaldi.
\newblock Learning multiple visual domains with residual adapters.
\newblock \emph{NeurIPS}, 30, 2017.

\bibitem[Reid et~al.(2024)Reid, Savinov, Teplyashin, Lepikhin, Lillicrap, Alayrac, Soricut, Lazaridou, Firat, Schrittwieser, et~al.]{reid2024gemini1_5}
Machel Reid, Nikolay Savinov, Denis Teplyashin, Dmitry Lepikhin, Timothy Lillicrap, Jean-baptiste Alayrac, Radu Soricut, Angeliki Lazaridou, Orhan Firat, Julian Schrittwieser, et~al.
\newblock Gemini 1.5: Unlocking multimodal understanding across millions of tokens of context.
\newblock \emph{arXiv preprint arXiv:2403.05530}, 2024.

\bibitem[Sharifymoghaddam et~al.(2024)Sharifymoghaddam, Upadhyay, Chen, and Lin]{sharifymoghaddam2024unirag}
Sahel Sharifymoghaddam, Shivani Upadhyay, Wenhu Chen, and Jimmy Lin.
\newblock Unirag: Universal retrieval augmentation for multi-modal large language models.
\newblock \emph{arXiv preprint arXiv:2405.10311}, 2024.

\bibitem[Shi et~al.(2024)Shi, Liu, Wang, Liao, Radhakrishnan, Huang, Yin, Sapra, Yacoob, Shi, et~al.]{shi2024eagle}
Min Shi, Fuxiao Liu, Shihao Wang, Shijia Liao, Subhashree Radhakrishnan, De-An Huang, Hongxu Yin, Karan Sapra, Yaser Yacoob, Humphrey Shi, et~al.
\newblock Eagle: Exploring the design space for multimodal llms with mixture of encoders.
\newblock \emph{arXiv preprint arXiv:2408.15998}, 2024.

\bibitem[Shi et~al.(2023)Shi, Min, Yasunaga, Seo, James, Lewis, Zettlemoyer, and Yih]{shi2023replug}
Weijia Shi, Sewon Min, Michihiro Yasunaga, Minjoon Seo, Rich James, Mike Lewis, Luke Zettlemoyer, and Wen-tau Yih.
\newblock Replug: Retrieval-augmented black-box language models.
\newblock \emph{arXiv preprint arXiv:2301.12652}, 2023.

\bibitem[Sidorov et~al.(2020)Sidorov, Hu, Rohrbach, and Singh]{sidorov2020textcaps}
Oleksii Sidorov, Ronghang Hu, Marcus Rohrbach, and Amanpreet Singh.
\newblock Textcaps: a dataset for image captioning with reading comprehension.
\newblock In \emph{ECCV}, pages 742--758, 2020.

\bibitem[Singh et~al.(2019)Singh, Natarajan, Shah, Jiang, Chen, Batra, Parikh, and Rohrbach]{singh2019textvqa}
Amanpreet Singh, Vivek Natarajan, Meet Shah, Yu Jiang, Xinlei Chen, Dhruv Batra, Devi Parikh, and Marcus Rohrbach.
\newblock Towards vqa models that can read.
\newblock In \emph{CVPR}, pages 8317--8326, 2019.

\bibitem[Su et~al.(2024)Su, Ahmed, Lu, Pan, Bo, and Liu]{su2024roformer}
Jianlin Su, Murtadha Ahmed, Yu Lu, Shengfeng Pan, Wen Bo, and Yunfeng Liu.
\newblock Roformer: Enhanced transformer with rotary position embedding.
\newblock \emph{Neurocomputing}, 568:\penalty0 127063, 2024.

\bibitem[Sun et~al.(2024)Sun, Yu, Cui, Zhang, Zhang, Wang, Gao, Liu, Huang, and Wang]{sun2023emu}
Quan Sun, Qiying Yu, Yufeng Cui, Fan Zhang, Xiaosong Zhang, Yueze Wang, Hongcheng Gao, Jingjing Liu, Tiejun Huang, and Xinlong Wang.
\newblock Generative pretraining in multimodality.
\newblock In \emph{ICLR}, 2024.

\bibitem[Sun et~al.(2022)Sun, Dong, Patra, Ma, Huang, Benhaim, Chaudhary, Song, and Wei]{sun2022length}
Yutao Sun, Li Dong, Barun Patra, Shuming Ma, Shaohan Huang, Alon Benhaim, Vishrav Chaudhary, Xia Song, and Furu Wei.
\newblock A length-extrapolatable transformer.
\newblock \emph{arXiv preprint arXiv:2212.10554}, 2022.

\bibitem[Taesiri(2024)]{arxivqa}
Mohammad~Reza Taesiri.
\newblock Arxivqa.
\newblock \url{https://github.com/taesiri/ArXivQA}, 2024.

\bibitem[Tanaka et~al.(2021)Tanaka, Nishida, and Yoshida]{tanaka2021visualmrc}
Ryota Tanaka, Kyosuke Nishida, and Sen Yoshida.
\newblock Visualmrc: Machine reading comprehension on document images.
\newblock In \emph{Proceedings of the AAAI Conference on Artificial Intelligence}, pages 13878--13888, 2021.

\bibitem[Tang et~al.(2023)Tang, Yang, Wang, Fang, Liu, Zhu, Zeng, Zhang, and Bansal]{tang2023unifying}
Zineng Tang, Ziyi Yang, Guoxin Wang, Yuwei Fang, Yang Liu, Chenguang Zhu, Michael Zeng, Cha Zhang, and Mohit Bansal.
\newblock Unifying vision, text, and layout for universal document processing.
\newblock In \emph{Proceedings of the IEEE/CVF conference on computer vision and pattern recognition}, pages 19254--19264, 2023.

\bibitem[Team et~al.(2023)Team, Anil, Borgeaud, Wu, Alayrac, Yu, Soricut, Schalkwyk, Dai, Hauth, et~al.]{team2023gemini}
Gemini Team, Rohan Anil, Sebastian Borgeaud, Yonghui Wu, Jean-Baptiste Alayrac, Jiahui Yu, Radu Soricut, Johan Schalkwyk, Andrew~M Dai, Anja Hauth, et~al.
\newblock Gemini: a family of highly capable multimodal models.
\newblock \emph{arXiv preprint arXiv:2312.11805}, 2023.

\bibitem[Team(2024{\natexlab{a}})]{InternVL2}
OpenGVLab Team.
\newblock Internvl2: Better than the best—expanding performance boundaries of open-source multimodal models with the progressive scaling strategy, 2024{\natexlab{a}}.

\bibitem[Team(2024{\natexlab{b}})]{qwen2.5}
Qwen Team.
\newblock Qwen2.5: A party of foundation models, 2024{\natexlab{b}}.

\bibitem[Tian et~al.(2024)Tian, Zhu, Xiong, Wang, Chen, Wang, Chen, Lu, Lu, Zhou, et~al.]{tian2024mminterleaved}
Changyao Tian, Xizhou Zhu, Yuwen Xiong, Weiyun Wang, Zhe Chen, Wenhai Wang, Yuntao Chen, Lewei Lu, Tong Lu, Jie Zhou, et~al.
\newblock Mm-interleaved: Interleaved image-text generative modeling via multi-modal feature synchronizer.
\newblock \emph{arXiv preprint arXiv:2401.10208}, 2024.

\bibitem[Tito et~al.(2023)Tito, Karatzas, and Valveny]{tito2023mpdocvqa}
Rub{\`e}n Tito, Dimosthenis Karatzas, and Ernest Valveny.
\newblock Hierarchical multimodal transformers for multipage docvqa.
\newblock \emph{Pattern Recognition}, 144:\penalty0 109834, 2023.

\bibitem[Tu et~al.(2024)Tu, He, Huang, Zhang, Yang, and Zhao]{tu2024overview}
Xiaoguang Tu, Zhi He, Yi Huang, Zhi-Hao Zhang, Ming Yang, and Jian Zhao.
\newblock An overview of large ai models and their applications.
\newblock \emph{Visual Intelligence}, 2\penalty0 (1):\penalty0 1--22, 2024.

\bibitem[Van~Landeghem et~al.(2023)Van~Landeghem, Tito, Borchmann, Pietruszka, Joziak, Powalski, Jurkiewicz, Coustaty, Ackaert, Valveny, et~al.]{van2023document}
Jordy Van~Landeghem, Ruben Tito, {\L}ukasz Borchmann, Micha{\l} Pietruszka, Pawe{\l} Joziak, Rafa{\l} Powalski, Dawid Jurkiewicz, Mickael Coustaty, Bertrand Ackaert, Ernest Valveny, et~al.
\newblock Document understanding dataset and evaluation (dude).
\newblock In \emph{Proceedings IEEE/CVF international conference on computer vision-ICCV 2023}, pages 19528--19540. IEEE/CVF, 2023.

\bibitem[Wang et~al.(2024{\natexlab{a}})Wang, Xu, Zhao, Ouyang, Wu, Zhao, Xu, Liu, Qu, Shang, et~al.]{wang2024mineru}
Bin Wang, Chao Xu, Xiaomeng Zhao, Linke Ouyang, Fan Wu, Zhiyuan Zhao, Rui Xu, Kaiwen Liu, Yuan Qu, Fukai Shang, et~al.
\newblock Mineru: An open-source solution for precise document content extraction.
\newblock \emph{arXiv preprint arXiv:2409.18839}, 2024{\natexlab{a}}.

\bibitem[Wang et~al.(2023{\natexlab{a}})Wang, Raman, Sibue, Ma, Babkin, Kaur, Pei, Nourbakhsh, and Liu]{wang2023docllm}
Dongsheng Wang, Natraj Raman, Mathieu Sibue, Zhiqiang Ma, Petr Babkin, Simerjot Kaur, Yulong Pei, Armineh Nourbakhsh, and Xiaomo Liu.
\newblock Docllm: A layout-aware generative language model for multimodal document understanding.
\newblock \emph{arXiv preprint arXiv:2401.00908}, 2023{\natexlab{a}}.

\bibitem[Wang et~al.(2024{\natexlab{b}})Wang, Bai, Tan, Wang, Fan, Bai, Chen, Liu, Wang, Ge, et~al.]{wang2024qwen2vl}
Peng Wang, Shuai Bai, Sinan Tan, Shijie Wang, Zhihao Fan, Jinze Bai, Keqin Chen, Xuejing Liu, Jialin Wang, Wenbin Ge, et~al.
\newblock Qwen2-vl: Enhancing vision-language model's perception of the world at any resolution.
\newblock \emph{arXiv preprint arXiv:2409.12191}, 2024{\natexlab{b}}.

\bibitem[Wang et~al.(2023{\natexlab{b}})Wang, Lv, Yu, Hong, Qi, Wang, Ji, Yang, Zhao, Song, et~al.]{wang2023cogvlm}
Weihan Wang, Qingsong Lv, Wenmeng Yu, Wenyi Hong, Ji Qi, Yan Wang, Junhui Ji, Zhuoyi Yang, Lei Zhao, Xixuan Song, et~al.
\newblock Cogvlm: Visual expert for pretrained language models.
\newblock \emph{arXiv preprint arXiv:2311.03079}, 2023{\natexlab{b}}.

\bibitem[Wang et~al.(2024{\natexlab{c}})Wang, Ren, Luo, Li, Yan, Chen, Wang, Li, Lu, Zhu, et~al.]{wang2024allseeingv2}
Weiyun Wang, Yiming Ren, Haowen Luo, Tiantong Li, Chenxiang Yan, Zhe Chen, Wenhai Wang, Qingyun Li, Lewei Lu, Xizhou Zhu, et~al.
\newblock The all-seeing project v2: Towards general relation comprehension of the open world.
\newblock \emph{arXiv preprint arXiv:2402.19474}, 2024{\natexlab{c}}.

\bibitem[Wang et~al.(2024{\natexlab{d}})Wang, Shi, Li, Wang, Huang, Xing, Chen, Li, Zhu, Cao, et~al.]{wang2023allseeing}
Weiyun Wang, Min Shi, Qingyun Li, Wenhai Wang, Zhenhang Huang, Linjie Xing, Zhe Chen, Hao Li, Xizhou Zhu, Zhiguo Cao, et~al.
\newblock The all-seeing project: Towards panoptic visual recognition and understanding of the open world.
\newblock In \emph{ICLR}, 2024{\natexlab{d}}.

\bibitem[Wang et~al.(2024{\natexlab{e}})Wang, Zhang, Ren, Duan, Li, Liu, Hu, Chen, Zhang, Lu, Zhu, Luo, Qiao, Dai, Shao, and Wang]{wang2024needle}
Weiyun Wang, Shuibo Zhang, Yiming Ren, Yuchen Duan, Tiantong Li, Shuo Liu, Mengkang Hu, Zhe Chen, Kaipeng Zhang, Lewei Lu, Xizhou Zhu, Ping Luo, Yu Qiao, Jifeng Dai, Wenqi Shao, and Wenhai Wang.
\newblock Needle in a multimodal haystack.
\newblock \emph{arXiv preprint arXiv:2406.07230}, 2024{\natexlab{e}}.

\bibitem[Wang et~al.(2024{\natexlab{f}})Wang, Zhang, Ren, Duan, Li, Liu, Hu, Chen, Zhang, Lu, et~al.]{wang2024mmniah}
Weiyun Wang, Shuibo Zhang, Yiming Ren, Yuchen Duan, Tiantong Li, Shuo Liu, Mengkang Hu, Zhe Chen, Kaipeng Zhang, Lewei Lu, et~al.
\newblock Needle in a multimodal haystack.
\newblock \emph{arXiv preprint arXiv:2406.07230}, 2024{\natexlab{f}}.

\bibitem[Wang et~al.(2023{\natexlab{c}})Wang, Zhou, Feng, Zhou, and Li]{wang2023towards}
Yonghui Wang, Wengang Zhou, Hao Feng, Keyi Zhou, and Houqiang Li.
\newblock Towards improving document understanding: An exploration on text-grounding via mllms.
\newblock \emph{arXiv preprint arXiv:2311.13194}, 2023{\natexlab{c}}.

\bibitem[Wei et~al.(2023)Wei, Kong, Chen, Zhao, Ge, Yang, Sun, Han, and Zhang]{wei2023vary}
Haoran Wei, Lingyu Kong, Jinyue Chen, Liang Zhao, Zheng Ge, Jinrong Yang, Jianjian Sun, Chunrui Han, and Xiangyu Zhang.
\newblock Vary: Scaling up the vision vocabulary for large vision-language models.
\newblock \emph{arXiv preprint arXiv:2312.06109}, 2023.

\bibitem[Xia et~al.(2024)Xia, Mao, Yan, Zhou, Zhang, Peng, Pi, Fu, Wu, Ye, et~al.]{xia2024docgenome}
Renqiu Xia, Song Mao, Xiangchao Yan, Hongbin Zhou, Bo Zhang, Haoyang Peng, Jiahao Pi, Daocheng Fu, Wenjie Wu, Hancheng Ye, et~al.
\newblock Docgenome: An open large-scale scientific document benchmark for training and testing multi-modal large language models.
\newblock \emph{arXiv preprint arXiv:2406.11633}, 2024.

\bibitem[Xiao et~al.(2023)Xiao, Tian, Chen, Han, and Lewis]{xiao2023efficient}
Guangxuan Xiao, Yuandong Tian, Beidi Chen, Song Han, and Mike Lewis.
\newblock Efficient streaming language models with attention sinks.
\newblock \emph{arXiv preprint arXiv:2309.17453}, 2023.

\bibitem[Yang(2023)]{yang2023longqlora}
Jianxin Yang.
\newblock Longqlora: Efficient and effective method to extend context length of large language models.
\newblock \emph{arXiv preprint arXiv:2311.04879}, 2023.

\bibitem[Yao et~al.(2024)Yao, Yu, Zhang, Wang, Cui, Zhu, Cai, Li, Zhao, He, et~al.]{yao2024minicpm_v}
Yuan Yao, Tianyu Yu, Ao Zhang, Chongyi Wang, Junbo Cui, Hongji Zhu, Tianchi Cai, Haoyu Li, Weilin Zhao, Zhihui He, et~al.
\newblock Minicpm-v: A gpt-4v level mllm on your phone.
\newblock \emph{arXiv preprint arXiv:2408.01800}, 2024.

\bibitem[Ye et~al.(2023{\natexlab{a}})Ye, Hu, Xu, Ye, Yan, Dan, Zhao, Xu, Li, Tian, et~al.]{ye2023mplugdocowl}
Jiabo Ye, Anwen Hu, Haiyang Xu, Qinghao Ye, Ming Yan, Yuhao Dan, Chenlin Zhao, Guohai Xu, Chenliang Li, Junfeng Tian, et~al.
\newblock mplug-docowl: Modularized multimodal large language model for document understanding.
\newblock \emph{arXiv preprint arXiv:2307.02499}, 2023{\natexlab{a}}.

\bibitem[Ye et~al.(2023{\natexlab{b}})Ye, Hu, Xu, Ye, Yan, Xu, Li, Tian, Qian, Zhang, et~al.]{ye2023ureader}
Jiabo Ye, Anwen Hu, Haiyang Xu, Qinghao Ye, Ming Yan, Guohai Xu, Chenliang Li, Junfeng Tian, Qi Qian, Ji Zhang, et~al.
\newblock Ureader: Universal ocr-free visually-situated language understanding with multimodal large language model.
\newblock \emph{arXiv preprint arXiv:2310.05126}, 2023{\natexlab{b}}.

\bibitem[Yu et~al.(2024)Yu, Tang, Xu, Cui, Ran, Yan, Liu, Wang, Han, Liu, et~al.]{yu2024visrag}
Shi Yu, Chaoyue Tang, Bokai Xu, Junbo Cui, Junhao Ran, Yukun Yan, Zhenghao Liu, Shuo Wang, Xu Han, Zhiyuan Liu, et~al.
\newblock Visrag: Vision-based retrieval-augmented generation on multi-modality documents.
\newblock \emph{arXiv preprint arXiv:2410.10594}, 2024.

\bibitem[Zhang et~al.(2024{\natexlab{a}})Zhang, Bai, Lv, Gu, Liu, Zou, Cao, Hou, Dong, Feng, et~al.]{zhang2024longcite}
Jiajie Zhang, Yushi Bai, Xin Lv, Wanjun Gu, Danqing Liu, Minhao Zou, Shulin Cao, Lei Hou, Yuxiao Dong, Ling Feng, et~al.
\newblock Longcite: Enabling llms to generate fine-grained citations in long-context qa.
\newblock \emph{arXiv e-prints}, pages arXiv--2409, 2024{\natexlab{a}}.

\bibitem[Zhang et~al.(2024{\natexlab{b}})Zhang, Hou, Lv, Cao, Hou, Niu, Hou, Dong, Feng, and Li]{zhang2024longreward}
Jiajie Zhang, Zhongni Hou, Xin Lv, Shulin Cao, Zhenyu Hou, Yilin Niu, Lei Hou, Yuxiao Dong, Ling Feng, and Juanzi Li.
\newblock Longreward: Improving long-context large language models with ai feedback.
\newblock \emph{arXiv preprint arXiv:2410.21252}, 2024{\natexlab{b}}.

\bibitem[Zhang et~al.(2024{\natexlab{c}})Zhang, Dong, Zang, Cao, Qian, Chen, Guo, Duan, Wang, Ouyang, et~al.]{zhang2024internlm}
Pan Zhang, Xiaoyi Dong, Yuhang Zang, Yuhang Cao, Rui Qian, Lin Chen, Qipeng Guo, Haodong Duan, Bin Wang, Linke Ouyang, et~al.
\newblock Internlm-xcomposer-2.5: A versatile large vision language model supporting long-contextual input and output.
\newblock \emph{arXiv preprint arXiv:2407.03320}, 2024{\natexlab{c}}.

\bibitem[Zhu et~al.(2024)Zhu, Hessel, Awadalla, Gadre, Dodge, Fang, Yu, Schmidt, Wang, and Choi]{zhu2024mmc4}
Wanrong Zhu, Jack Hessel, Anas Awadalla, Samir~Yitzhak Gadre, Jesse Dodge, Alex Fang, Youngjae Yu, Ludwig Schmidt, William~Yang Wang, and Yejin Choi.
\newblock Multimodal c4: An open, billion-scale corpus of images interleaved with text.
\newblock \emph{NIPS}, 36, 2024.

\end{thebibliography}
}

\clearpage
\setcounter{page}{1}
\maketitlesupplementary

\section{Details of Data Construction}
\subsection{Document Data Format}
\begin{figure*}[t]
    \centering
    {\includegraphics[width=\linewidth]{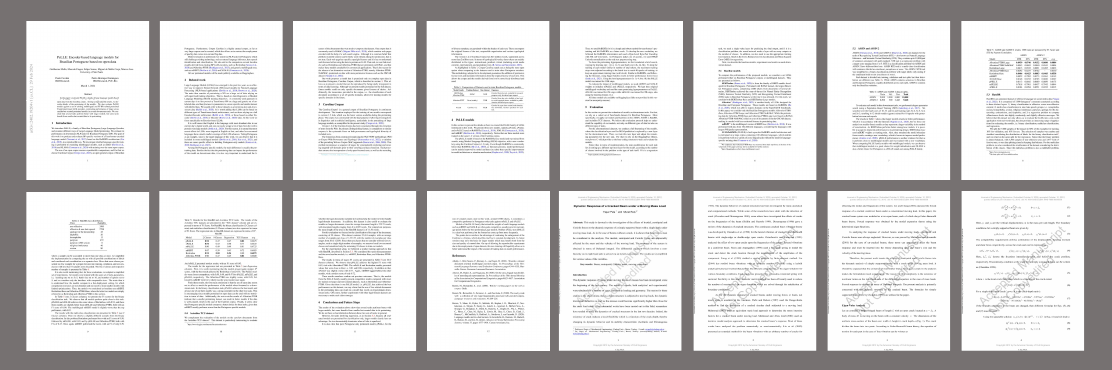}}
    \caption{
    \textbf{An example of multi-image format document.} Each page of the document is rendered as an image.
    }
    \label{fig:multi-image}
\end{figure*}

\begin{figure*}[t]
    \centering
    \small
    {\includegraphics[width=.95\linewidth]{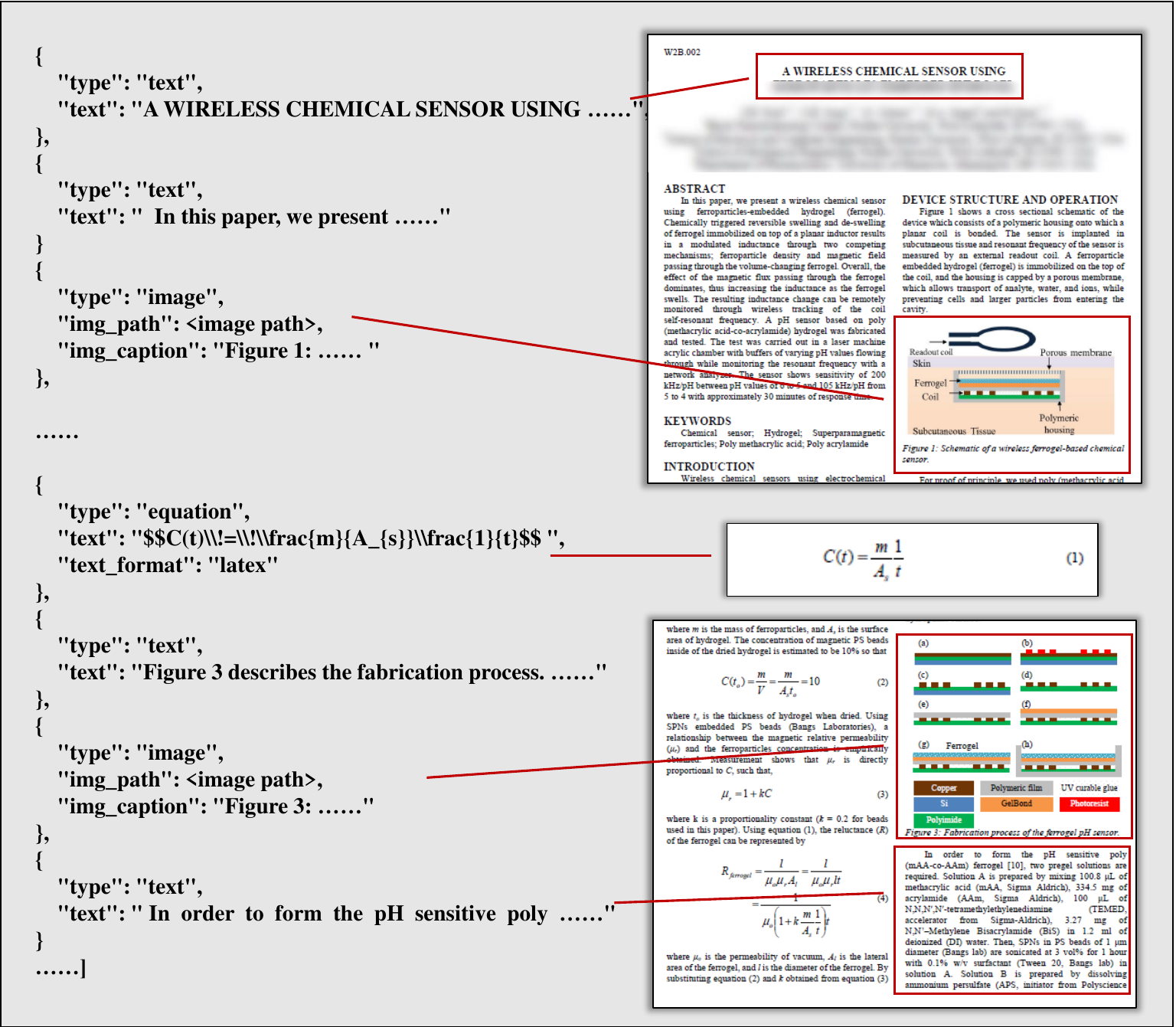}}
    \caption{
    \textbf{An example of interleaved text-image format document.} We capture the documents' textual content and construct them into interleaved images and text. 
    }
    \label{fig:interleaved-image_text}
\end{figure*}

As stated in Section~\ref{sec:data_engine}, documents in Doc-750K can be extracted in two formats: \textbf{Interleaved Text-Image Format} and rendered \textbf{Multi-Image Format}. The interleaved format utilizes an external PDF parser to extract text directly, avoiding OCR errors. However, this approach sacrifices some layout information. The multi-image format, commonly used in screenshot-based QA scenarios, preserves the complete layout but relies on the model's built-in OCR capabilities, which may introduce errors. Each format has its advantages and is suitable for different applications. By leveraging both formats, the model can develop complementary capabilities, enhancing its robustness across diverse input types. Examples of each format are shown in Figure~\ref{fig:multi-image} and Figure~\ref{fig:interleaved-image_text}, respectively.

\begin{table*}[t]
\centering
\renewcommand\arraystretch{0.2}
\begin{tabular}{l|c|l|l}
\toprule
Datasets          & \#QA  & Image Types                          & Tasks \\ 
\midrule
MP-DocVQA~\cite{tito2023mpdocvqa}         & 46K   & PDF Documents 
& VQA \\
\midrule
DUDE~\cite{van2023document}              & 41K   & PDF Documents 
& \makecell[tl]{VQA} \\
\midrule
MP-DocStruct1M~\cite{hu2024mplugdocowl2}    & 1M    & PDF Documents 
& \makecell[tl]{Text Parsing, Text Lookup} \\
\midrule
MP-DocReason51K~\cite{hu2024mplugdocowl2}   & 51K   & \makecell[tl]{PDF Documents, \\ Infographics, Webpages, \\ Charts, Natural images}
& VQA \\
\midrule
DocGenome~\cite{xia2024docgenome}         & N/A   & \makecell[tl]{PDF Documents}
& \makecell[tl]{Layout Detection, Document Transformation} \\
\midrule
\dataname (ours)  & 758K  & \makecell[tl]{PDF documents, \\ Charts, Tables} 
& \makecell[tl]{VQA, Abstract Writing, Paper Titling, \\ Caption Writing, Experiment Writing, \\ Translation, Review, Reply }    \\ 
\bottomrule
\end{tabular}
\caption{\textbf{Comparison with other document-level datasets. }}
\label{tab:addi_dataset_comparison}
\end{table*}

\subsection{Image Types and Tasks Coverage}
We provide additional comparisons of \dataname with various document-level datasets in terms of image types and task coverage, as shown in Table~\ref{tab:addi_dataset_comparison}. Compared to these datasets, \dataname exhibits greater diversity in proxy tasks and ranks among the largest datasets in terms of QA pair count.

\subsection{Question-Answer Pairs Generation}
\noindent\textbf{Prompt.} The prompt used to generate question-answer pairs from GPT-4o is shown below.

\promptbox{Please read the paper and first check if this is an English paper. If it is not an English paper, don't do any other things. If it is an English paper, please design about 3 to 5 question-answer pairs based on the paper. All questions should require as much text as possible to answer and it is better to ask about the images in the papers. All images in the questions should be represented as the mentioned title in the paper like Figure 1/Figure 2 or Table 1/ Table 2 and they must be mentioned in the questions. The question should be specific enough that it can only be answered with the paper. The question should also be interesting and intellectual enough that a curious reader of the paper would ask about it. The answer of the QA pair must start with "According to the original text ......”, first give the relevant original text in the reference content, and then answer the question in detail. Please try to analyze the asked image in the answer. Please directly output a list of the QA pairs without any other outputs. Here is the paper: \\
<paper>
}

\noindent\textbf{Examples.} In Section~\ref{sec:data_engine}, we propose diverse question-answer formats tailored to data from different sources. To fully utilize the webpage structure of OpenReview, we develop tasks focused on review writing and replies within its review-reply framework. For Sci-Hub and Arxiv, we use their well-defined writing structures to create tasks such as writing and translating various sections. We provide examples of these various QA formats in Figure~\ref{fig:qa_example}.  

\noindent\textbf{Quality Evaluation.} The quality of \dataname is ensured through the following measures: 
(1) Human-Originated Data: QA pairs from OpenReview are derived from human-written discussions, providing high contextual quality.
(2) Structured Tasks: Tasks like abstract writing and paper titling are constructed based on document metadata, following deterministic rules to ensure reliability.
(3) Synthetic QA: We randomly sample and manually review 500 
training QA pairs across tasks and 498 of 500 (over 99\%) of the pairs were relevant. 

\begin{figure*}
  \centering
  \small
  \fbox{
    \parbox{.95\linewidth}{\texttt{\textbf{Review Writing} \\
    Question: Please review the following paper and provide a constructive critique. Focus on the methodology, results, and overall contributions, and highlight both strengths and areas for improvement. Your review should be detailed and insightful, offering suggestions for enhancing the research. Here is the paper: \\
    <paper>\\
    Answer: <review> \\ 
    \textbf{Reply Writing} \\
    Question: Given the following paper and its review, write a reply to address the feedback provided. Here is the paper:\\
    <paper>\\
    Here is the review:\\
    <review>\\
    Answer: <reply>\\ 
    \textbf{Abstract Writing} \\
    Question: Please read the full text of the following research paper and provide a concise summary in the form of an abstract. The summary should capture the main objectives, methods, results, and conclusions of the paper. Ensure that the abstract is clear, coherent, and informative for readers who have not read the full paper. Here is the paper:\\
    <paper w/o abstract>\\
    Answer: <abstract>\\
    \textbf{Paper Titling} \\
    Question: Based on the provided abstract or introduction of the research paper, please generate a concise and informative title that accurately reflects the main focus and contributions of the paper. The title should be engaging and clearly convey the essence of the research. Here is the abstract and introduction:\\
    <abstract> \\
    <introduction>\\
    Answer: <title>\\ 
    \textbf{Caption Writing} \\
    Question: The images or tables and their relative texts in a research paper are given interleaved as follows. Please write a caption for each image or table based on the relative texts provided. Here is the image:\\
    <image>\\
    Here is the relative text:\\
    <text>\\
    Answer: <caption>\\ 
    \textbf{Experiments Writing} \\
    Question: Please write the "Experiments" section based on the incomplete research paper provided. Ensure that the section is well-structured and includes the details of the Figures and tables. Here is the paper:\\
    <paper w/o experiments>\\
    Answer: <experiments section>\\ 
    \textbf{Translation} \\
    Question: Please read the full text of the following research paper and translate the Experiments/Abstract section into Chinese. Here is the paper:\\
    <paper> (Multi-Image format) \\
    Answer: <translation>\\ 
    \textbf{Multi-Turn QA (interleaved and rendered)} \\
    Question: Please read the full paper and answer the question. Here is the paper:\\
    <paper>\\
    Here is the question: <question> (generated)\\
    Answer: <answer> (generated)\\
    Question: <question> (generated)\\
    Answer: <answer> (generated)
    }}
  }
  \caption{\textbf{The examples of different formats of the QA pairs.} All tasks leverage the inherent structure of the documents.}
  \label{fig:qa_example}
\end{figure*}

\begin{table*}[t]
\centering
\setlength\tabcolsep{4pt}
\begin{tabular}{c|l|l|l}
\toprule
Layout    & Benchmark       & Description                             & Metric                                                                                      \\
\midrule
\multirow{3}{*}{Single-Page} & DocVQA~\cite{mathew2021docvqa}          & VQA on documents.                       & ANLS                                          \\
                             & ChartVQA~\cite{masry2022chartqa}        & VQA on charts.                          & Relaxed EM                                    \\
                             & InfoVQA~\cite{mathew2022infographicvqa}         & VQA on infographics.                    & ANLS                                          \\
\midrule
\multirow{5}{*}{Multi-Page}  & MP-DocVQA~\cite{tito2023mpdocvqa}       & VQA on multi-page documents.            & ANLS                                          \\
                             & MMLongBench-Doc~\cite{ma2024mmlong} & VQA on super-long PDF documents.        & Accuracy, F1 Score                            \\
                             & \multirow{3}{*}[3ex]{DocGenome~\cite{xia2024docgenome}}  & \multirow{3}{*}[3ex]{VQA on multi-page scientific documents.} & Classification Acc, \\
                             &                                  &                                                          & Title ED, Abstract ED, \\
                             &                                  &                                                          & Single-Page Acc,   Multi-Page Acc \\
\midrule
Interleaved                  & MM-NIAH~\cite{wang2024mmniah}         & VQA on natural texts or images.         & Accuracy   \\ 
\bottomrule
\end{tabular}
\caption{\textbf{Evaluation benchmarks and metrics for document understanding.}
This table summarizes key benchmarks used in document understanding tasks across three layout types: single-page, multi-page, and interleaved. 
}
\label{tab:benchmark_metrics}
\end{table*}

\section{Evaluation Details}

\subsection{Benchmark Metrics}

We report the metrics of benchmarks used in the evaluation in Table~\ref{tab:benchmark_metrics}. 
For DocVQA~\cite{mathew2021docvqa}, InfoVQA~\cite{mathew2022infographicvqa}, and MP-DocVQA~\cite{tito2023mpdocvqa}, we employ ANLS to evaluate the similarity between model responses and ground truth answers, while ChartQA~\cite{masry2022chartqa} uses Relaxed Exact Match (Relaxed EM). For open-ended QA tasks in MMLongbench-Doc~\cite{ma2024mmlong} and DocGenome~\cite{xia2024docgenome}, we utilize GPT-4o to assess the correctness of answers and calculate GPT Accuracy. Other tasks in DocGenome follow their official evaluation metrics.

\subsection{Evaluation Settings}

\textbf{For single-page benchmarks} such as DocVQA, ChartQA, and InfoVQA, we conduct the evaluation using a unified prompt as follows:

\promptbox{<image> \\ 
<question> \\ 
Answer the question using a single word or phrase.}

\noindent\textbf{For multi-page benchmarks}, we discuss them case by case. We employed image concatenation for multi-page VQA benchmarks like MP-DocVQA, MMLongBench-Doc, and DocGenome to reduce the excessive input patches. Adjacent pages were vertically concatenated into a single image, with a maximum total image count limit of 18.

(1) For MPDocVQA, we use the prompt for a $N$ concatenated page document as follows:

\promptbox{Image-1: <concat-page 1> \\ 
Image-2: <concat-page 2> \\ 
... \\ 
Image-N: <concat-page N> \\ 
<question> \\
Answer the question using a single word or phrase.
}

(2) For MMLongBench-Doc and DocGenome, we use the official prompt in their open-sourced code base for response generation and extract the correctness of the answer using GPT-4o.

(3) For MM-NIAH, we use the original interleaved data format and calculate the accuracy by their official judgment function.

\noindent\textbf{Evaluation of LLMs on Multimodal Document Benchmarks.}
We utilized the InternVL2-8B as the OCR model to extract text from each image of the document, followed by post-processing to remove redundant responses. The text extraction prompt is as follows:

\promptbox{Image-1: <image> \\
Please extract the text from Image-1, while retaining as much of the original formatting and structured information (such as headings, paragraphs, lists, tables, charts, etc.) as possible. If the document is not in PDF format, provide a caption for Image-1. Present the extracted information directly without additional explanations.
}

We concatenated the extracted texts to replace the original document images for the language models:
\vspace{-2ex}
\promptbox{Page 1: \\
<text 1>  \\
Page 2: \\
<text 2> \\
... \\
Page N: \\
<text N>
}

For the image-text interleaved data, we replaced the images with the captions as the input.

\noindent\textbf{Implementation Details of Multimodel RAG.}
VisRAG~\cite{yu2024visrag} uses their proposed retrieval model VisRet to calculate scores for each image and text segment based on the query. InternVL-RAG~\cite{wang2024needle} utilizes InternVL-14B, a CLIP-like model, to compute the similarity between images and text. For multi-paged VQA benchmarks, we select the top-3 retrieved documents for generation. For interleaved VQA, we choose up to 8K tokens for the generation.

\section{Training Details}
\subsection{Hyperparameters}
We report the models and training hyperparameters of \modelname-2B and \modelname-8B in Table~\ref{tab:hyperparam}.

\begin{table}[h]
\centering
\small
\setlength\tabcolsep{3.6pt}
\begin{tabular}{c|l|cc}
\toprule
\multicolumn{2}{c|}{Settings}              & \modelname-2B     & \modelname-8B             \\
\midrule
\multirow{2}{*}{\rotatebox{90}{Model}}     & ViT               & InternViT-300M      & InternViT-300M \\
                                           & LLM    & InternLM2-1.8B & InternLM2.5-7B \\

\midrule
\multirow{11}{*}{\rotatebox{90}{Training Hyperparameters}} & Tile Resolution   & \multicolumn{2}{c}{448}                                  \\
                                           & Batch Size        & \multicolumn{2}{c}{128}                                  \\
                                           & Optimizer         & \multicolumn{2}{c}{AdamW}                                \\
                                           & Learning Rate     & \multicolumn{2}{c}{1.00E-05}                             \\
                                           & Warmup Ratio      & \multicolumn{2}{c}{0.03}                                   \\
                                           & LR Scheduler      & \multicolumn{2}{c}{Cosine}                               \\
                                           & Weight Decay      & 0.01                & 0.05           \\
                                           & ViT Drop Path     & \multicolumn{2}{c}{0.1}                                  \\
                                           & Max Tile Number   & \multicolumn{2}{c}{24}                                   \\
                                           & Image Threshold   & \multicolumn{2}{c}{48}                                   \\
                                           & Token Threshold    & \multicolumn{2}{c}{32K}                                  \\
                                           & Epochs   & \multicolumn{2}{c}{1}              \\
\bottomrule
\end{tabular}
\caption{\textbf{Training settings and hyperparameters for \modelname models.} Key configurations for \modelname-2B and \modelname-8B, including model architectures and training parameters.}
\label{tab:hyperparam}
\end{table}

\subsection{Multimodal Packed Dataset}

In this section, we provide a detailed description of our packing algorithm. The main workflow is outlined in Algorithm \ref{alg:packed_dataset}. Specifically, our algorithm constructs the packed dataset by combining individual samples drawn from the original dataset. The packing operation involves four steps:

(1) \textbf{Check Sample:} Given an individual sample, we first verify whether the number of images exceeds the image threshold $T_i$ or the number of tokens exceeds the token threshold $T_t$. If either condition is met, the sample is truncated into $N$ parts. The first $N-1$ parts contain exactly $T_i$ images or $T_t$ tokens and are immediately added to the output. The remaining part is passed to the subsequent steps for further processing.

(2) \textbf{Find Buffer:} For the remaining part of the sample, we attempt to find a suitable buffer from the buffer list to pack it together with the sample. The combined result must not exceed the thresholds $T_i$ for images or $T_t$ for tokens, while maximizing the total number of images and tokens in the packed sample. To speed up this process, the buffer list is organized as a priority queue.

(3) \textbf{Pack Samples:} The given sample and the selected buffer are packed together. Notably, during the training process, each token can only attend to other tokens within the same original sample. Tokens from other samples packed together remain inaccessible.

(4) \textbf{Maintain Buffer List:} After generating a packed sample, we check if its number of images or tokens meets the specified thresholds. If so, the sample is added to the output; otherwise, it is reinserted into the buffer list for potential future packing.
Note that we omit numerous edge cases for brevity.

\begin{algorithm}
	\caption{Multimodal Packed Dataset} 
	\label{alg:packed_dataset} 

    \KwIn{Dataset $\mathcal{D}$, buffer list $\mathcal{B}$, Token Threshold $T_t$, Image Threshold $T_i$}
    \KwOut{Packed Dataset $\mathcal{D}_\text{packed}$}

    \ForEach{data\_sample $d$ in $\mathcal{D}$}{

        $b  \gets \texttt{find\_buffer}(d, B)$ 

        $b_{p} \gets \texttt{pack}(d, b)$

        \If{$b_{p}$ \text{contains more than} $T_i$ \text{images or} $T_t$ \text{tokens}}{
            \texttt{yield} $b_{packed}$
        }
        \Else{
            $\texttt{insert}(b_{packed}, B)$
        }
    }

\end{algorithm}

\section{Qualitative Examples}

In this section, we show a series of qualitative examples to illustrate the effectiveness of our \modelname in handling complex multi-page documents. Each figure highlights a specific capability of the model in addressing various tasks.

\begin{figure*}[]
    \centering
    {\includegraphics[width=0.65\linewidth]{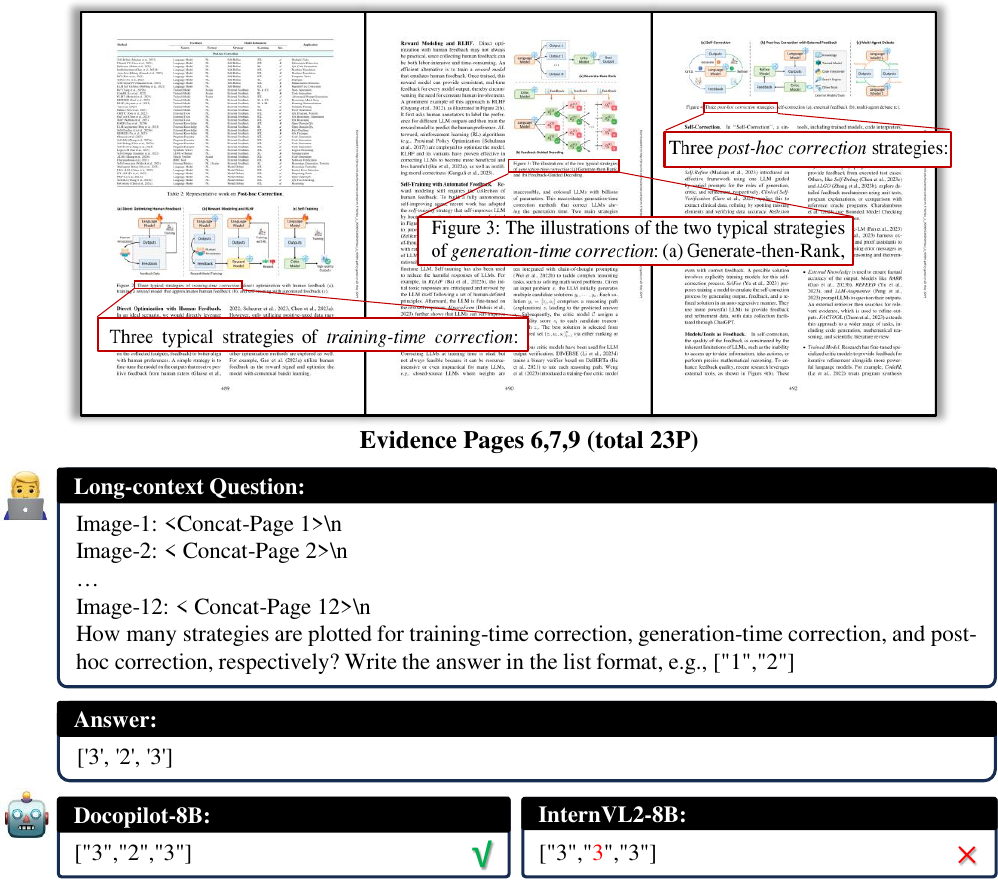}}
    \caption{\textbf{A qualitative example of retrieval in a multi-page document.}}
    \label{fig:good_sample1}
\end{figure*}

Figure~\ref{fig:good_sample1} demonstrates the model's ability to accurately retrieve relevant information from a multi-page document, showcasing its capability to perform robust cross-page retrieval tasks.

\begin{figure*}[]
    \centering
    {\includegraphics[width=0.65\linewidth]{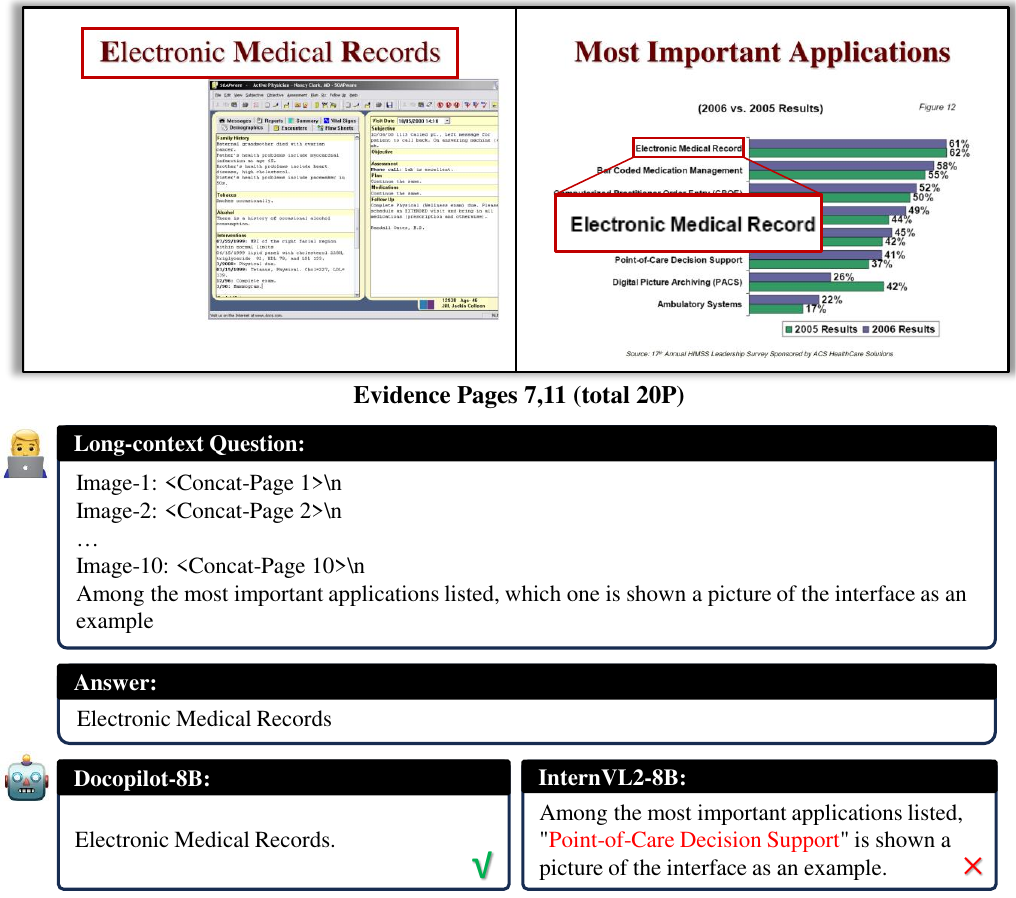}}
    \caption{\textbf{A qualitative example of backward query in a multi-page document.}}
    \label{fig:good_sample2}
\end{figure*}

In Figure~\ref{fig:good_sample2}, we illustrate the model's proficiency in performing backward queries, where context must be traced across pages in reverse order to locate relevant content.

\begin{figure*}[]
    \centering
    {\includegraphics[width=0.6\linewidth]{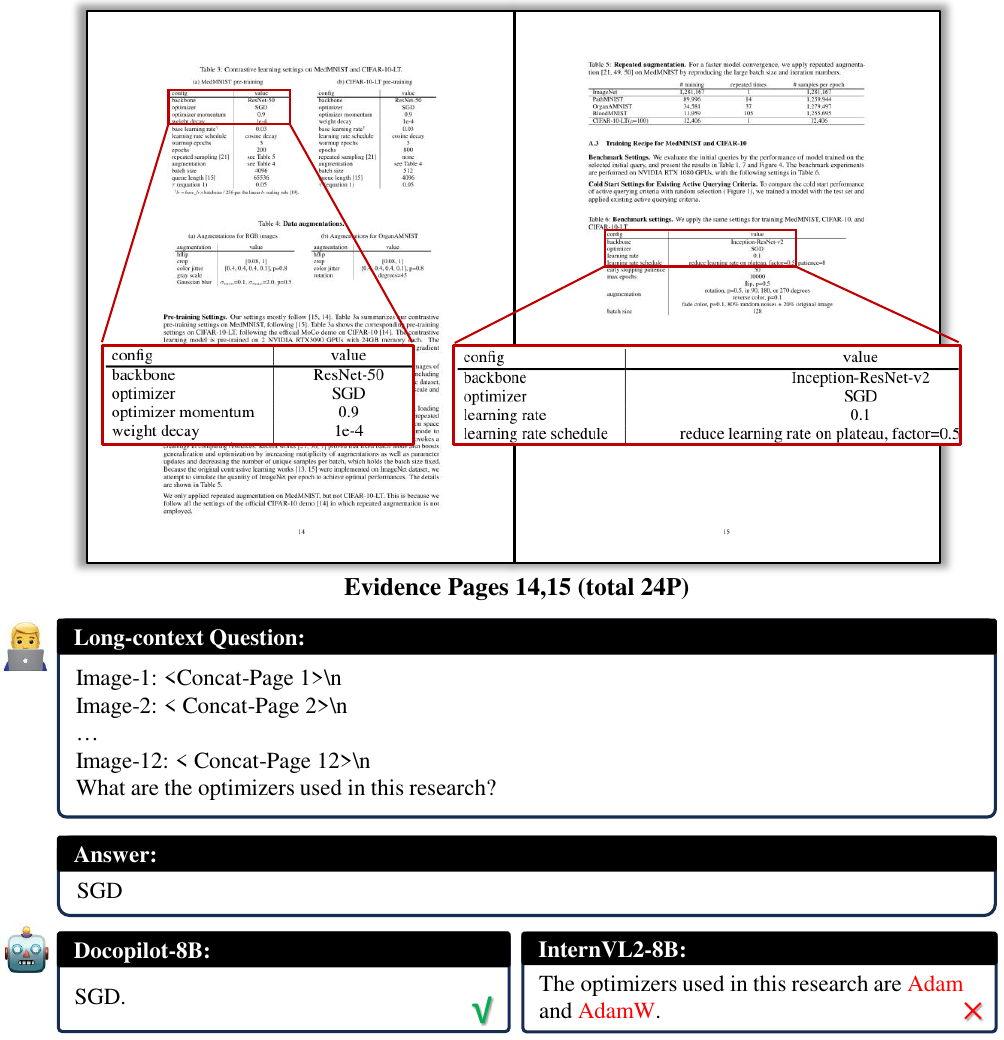}}
    \caption{\textbf{A qualitative example of consistency of answers in multi pages.}}
    \label{fig:good_sample3}
\end{figure*}

Figure~\ref{fig:good_sample3} highlights the consistency of answers when the model is queried across multiple pages. This example demonstrates that the model maintains coherence and accuracy even when information is distributed across different parts of the document.

\begin{figure*}[]
    \centering
    {\includegraphics[width=0.6\linewidth]{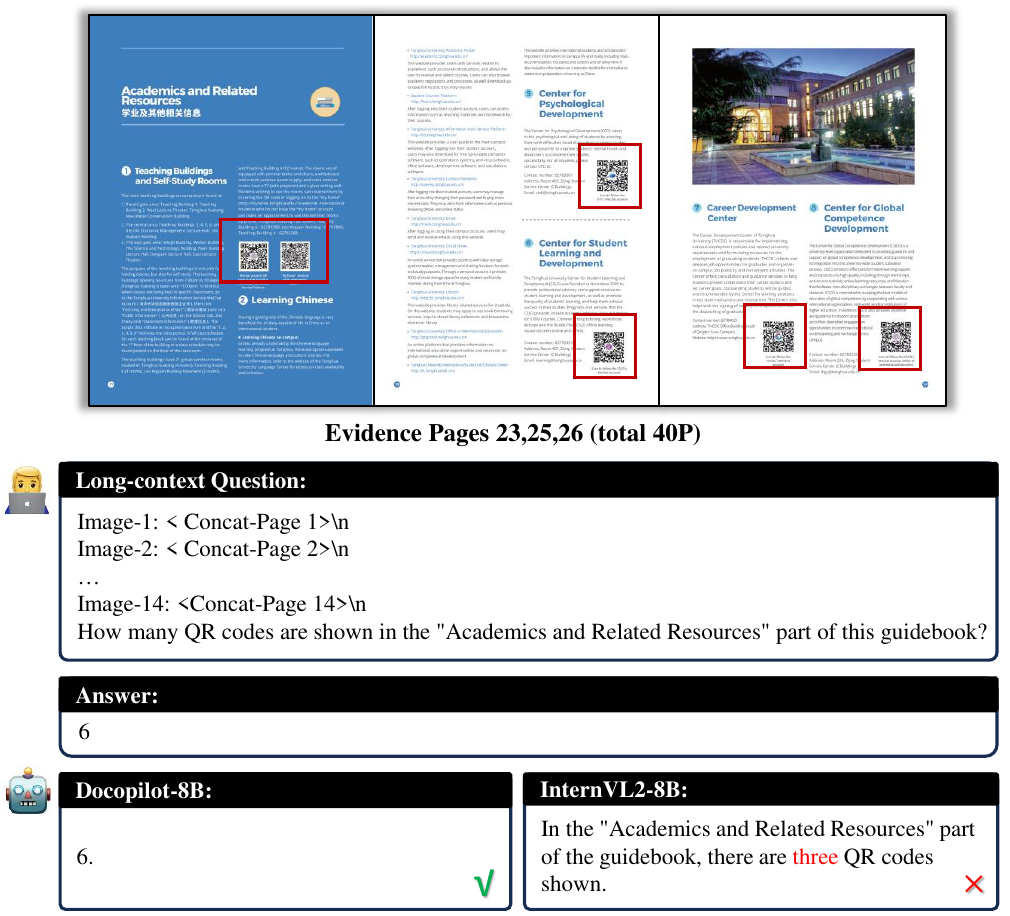}}
    \caption{\textbf{A qualitative example of counting across pages.}}
    \label{fig:good_sample4}
\end{figure*}

Figure~\ref{fig:good_sample4} presents an example of counting across pages, showcasing the model's ability to integrate numerical information from disparate locations in a document. 

\begin{figure*}[t]
    \centering
    {\includegraphics[width=0.6\linewidth]{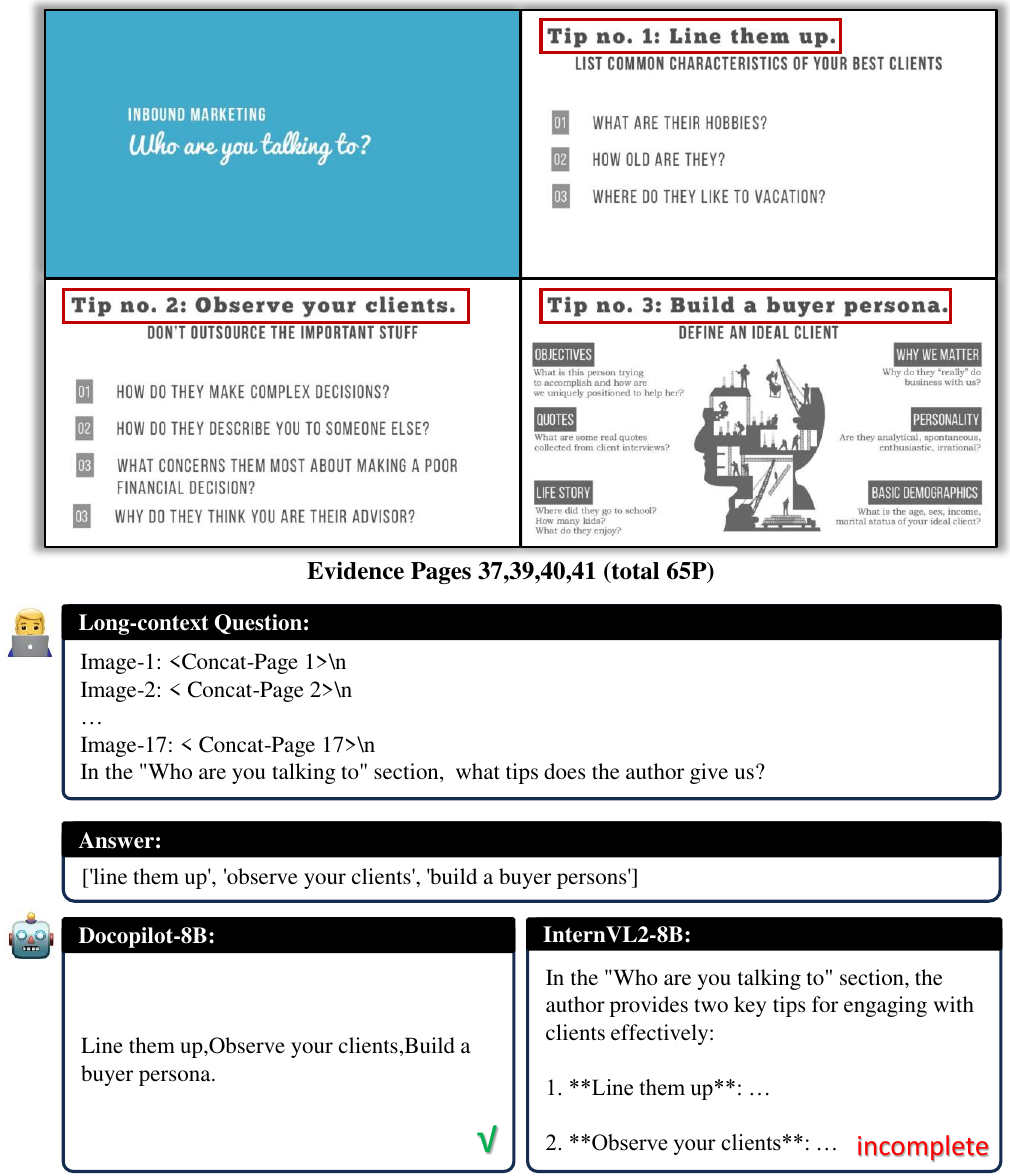}}
    \caption{\textbf{A qualitative example of retrieval in a specific section of a super-long document.}}
    \label{fig:good_sample5}
\end{figure*}

Lastly, Figure~\ref{fig:good_sample5} demonstrates the model's capability to pinpoint specific information within a designated section of a super-long document, emphasizing its fine-grained retrieval capabilities.

These examples collectively highlight the robustness and adaptability of our approach in understanding and processing multi-page documents effectively.

\section{Limitations}
The objective of \dataname is to develop a large-scale, multi-task, multimodal document-level QA dataset that efficiently trains MLLMs for document understanding. Sourced primarily from open academic platforms, the dataset focuses on tasks like multi-page QA, reasoning, and translation, with some academic-specific tasks such as titling and summarization. 
A key limitation of \dataname is its current domain restriction to academic documents. We plan to expand the dataset’s coverage to a broader range of document types and enhance the generalizability of proxy tasks, ensuring wider applicability across diverse domains.

\end{CJK}
\end{document}